\documentclass[lettersize,journal]{IEEEtran}
\usepackage{amsmath,amsfonts}
\usepackage{algorithm}
\usepackage{algpseudocode}
\usepackage{array}
\usepackage[caption=false,font=footnotesize,labelfont=rm,textfont=rm]{subfig}
\usepackage{textcomp}
\usepackage{stfloats}
\usepackage{url}
\usepackage{verbatim}
\usepackage{graphicx}
\usepackage{cite}
\usepackage{tikz}   % 正文矩形框
\usepackage[breaklinks, colorlinks, linkcolor=blue, citecolor=blue, urlcolor=blue]{hyperref}
\usepackage{simpleicons}
\usepackage{pifont}
\usepackage{orcidlink}
\usepackage{listings}

\usepackage{booktabs}
\usepackage{amssymb}
\usepackage{xcolor}
\usepackage{multirow}

\usepackage{color, colortbl}
\definecolor{Gray}{gray}{0.9}

\newcommand\dunderline[2][.4pt]{%
  \raisebox{-#1}{\underline{\raisebox{#1}{\smash{\underline{#2}}}}}}

\newcolumntype{S}{>{\footnotesize}c}

\newcommand{\NMSAP}{
\begin{table}[!t]
\centering
\caption{Effect of rotated IoU thresholds and confidence thresholds on RTMDet-s accuracy for DOTA-v1.0.}
\begin{minipage}{0.49\linewidth}
\centering
    \setlength{\tabcolsep}{3pt}
    \begin{tabular}{c|ccc}
    \toprule
    \textbf{RIoU thr.}      &  \multirow{2}{*}{0.05}  & \multirow{2}{*}{0.3} & \multirow{2}{*}{0.5} \\ 
    \textbf{(Conf=0.05)}    &                         &                      &  \\
    \midrule
    \textbf{AP$_{50}$ (\%)} & 76.53                 & 76.72                 & 76.70\\
    \bottomrule
    \end{tabular}
\end{minipage}
\begin{minipage}{0.49\linewidth}
\centering
    \setlength{\tabcolsep}{3pt}
    \begin{tabular}{c|ccc}
    \toprule
    \textbf{Conf thr.}   &  \multirow{2}{*}{0.01}   & \multirow{2}{*}{0.05}   & \multirow{2}{*}{0.25}       \\
    \textbf{(RIoU=0.1)}  &                      &     \\
    \midrule
    \textbf{AP$_{50}$ (\%)} & 76.82                & 76.86                 & 75.61\\
    \bottomrule
    \end{tabular}
\end{minipage}
\vspace{-0.5em}
\label{NMSAP}
\end{table}
}

\newcommand{\TableExpSetting}{
    \begin{table}[t]
    \centering
    \caption{Main hyperparameter of our detectors.}
    \resizebox{\linewidth}{!}
    {  
        \begin{tabular}{l|ccc|ccc}
        \toprule
        \multirow{2}{*}{\textbf{Setting}} & \multicolumn{3}{c|}{\textbf{O$^2$-RTDETR}} & \multicolumn{3}{c}{\textbf{O$^2$-DFINE}}   \\
                                 & \textbf{R18} & \textbf{R34} & \textbf{R50} & \textbf{S} & \textbf{M} & \textbf{L} \\
        \midrule
        Optimizer                & AdamW    & AdamW   & AdamW   & AdamW   & AdamW   & AdamW    \\
        Base LR                  & 1e-4     & 1e-4    & 1e-4    & 5e-5    & 5e-5    & 6.25e-5  \\
        Backbone LR              & 1e-5     & 1e-5    & 1e-5    & 2.5e-6  & 5e-6    & 3.125e-5 \\
        Weight decay             & 1e-4     & 1e-4    & 1e-4    & 1e-4    & 1e-4    & 1.25e-4  \\
        Clip grad                & 0.1      & 0.1     & 0.1     & 0.1     & 0.1     & 0.1      \\
        Queries $K$              & 300      & 300     & 300     & 300     & 300     & 300      \\
        Layers $L$               & 6        & 6       & 6       & 3       & 4       & 6        \\
        Batchsize                & 8        & 8       & 8       & 8       & 8       & 8        \\
        Epochs                   & 72       & 72      & 72      & 72      & 72      & 72       \\
        \bottomrule
        \end{tabular}
    }
            \label{expsetting}
    \end{table}
}

\newcommand{\DIORRResults}{
\begin{table*}[!htbp]
    \setlength{\tabcolsep}{4pt}
    \caption{Performance comparisons on the \textbf{DIOR-R} dataset. The \textbf{best} and \dunderline{second-best} results are highlighted in \textbf{bold} and \dunderline{underlined}.}
    \resizebox{\linewidth}{!}
    {
    \begin{tabular}{lccccccccccccccccccccccc}
    \toprule
    \textbf{Method} & \textbf{Back.} & \textbf{\#P} & \textbf{APL} & \textbf{APO} & \textbf{BF} & \textbf{BC} & \textbf{BR} & \textbf{CH} & \textbf{DAM} & \textbf{ETS} & \textbf{ESA} & \textbf{GF} & \textbf{GTF} & \textbf{HA} & \textbf{OP} & \textbf{SH} & \textbf{STA} & \textbf{STO} & \textbf{TC} & \textbf{TS} & \textbf{VE} & \textbf{WM} & \textbf{AP$_{50}$}  \\
    \midrule
    RoI Trans.~\cite{roi_transformer} & R50 & 55M & 63.34    & 37.88    & 71.78    & 87.53    & 40.68    & 72.60    & 26.86    & \dunderline{78.71}    & 68.09    & 68.96    & 82.74    & 47.71    & 55.61    & 81.21    & \dunderline{78.23} & 70.26    & 81.61    & 54.86    & 43.27    & 65.52    & 63.87    \\
    O-RCNN~\cite{orientedrcnn} & R50 & 41M & 71.10 & 39.30 & \textbf{79.50} & 86.20 & 43.10 & 72.60 & 29.50 & 66.70 & 79.30 & 68.60 & 82.40 & 43.60 & 57.60 & 81.30 & 74.20 & 62.60 & 81.40 & 54.80 & 46.80 & 66.00 & 64.30 \\
    AOPG~\cite{dior} & R50  & - & 62.39    & 37.79    & 71.62    & 87.63    & 40.90    & 72.47    & 31.08    & 65.42    & 77.99    & 73.20    & 81.94    & 42.32    & 54.45    & 81.17    & 72.69    & 71.31    & 81.49    & 60.04 & 52.38    & 69.99    & 64.41    \\
    ARS-DETR~\cite{arsdetr} & R50  & 41M & 68.00    & 54.17    & 74.43    & 81.65    & 41.13    & 75.66    & 34.89    & 73.07    & 81.92 & 76.10 & 78.62  & 36.33    & 55.41    & 84.55    & 70.09    & 72.23    & 81.14    & 61.52    & 50.57    & 70.28    & 66.12    \\
    RO$^2$-DETR~\cite{ro2detr} & R50 & - &  69.78 & 54.77 & 74.49 & 82.11 & 42.30 & 76.66 & 33.94 & 72.99 & 82.01 & 76.11 & 78.56 & 40.11 & 56.06 & 85.06 & 73.11 & 72.19 & 84.10 & 60.66 & 53.02 & 71.11 & 66.43 \\
    OrientedRep~\cite{oriented_reppoints} & R50 & 36M & 70.03 & 46.11    & 76.12    & 87.19    & 39.14    & 78.76    & 34.57    & 71.80    & 80.42    & 76.16    & 79.41    & 45.48    & 54.90    & 87.82    & 77.03    & 68.07    & 81.60    & 56.83    & 51.57    & 71.25 & 66.71    \\
    DCFL~\cite{dcfl} & R50  & 36M & 68.60    & 53.10    & 76.70 & 87.10    & 42.10    & 78.60 & 34.50    & 71.50    & 80.80    & \textbf{79.70} & 79.50    & 47.30    & 57.40    & 85.20    & 64.60    & 66.40    & 81.50    & 58.90    & 50.90    & 70.90    & 66.80    \\
    OFormer~\cite{orientedformer} & R50  & 44M & 65.65    & 48.69 & 78.79    & 87.17    & 41.90 & 76.34    & 34.37 & 72.14    & 81.40    & 75.34    & 79.83    & 45.15    & 56.12 & \dunderline{88.66} & 67.59  & 72.68 & 87.32    & 60.31    & 56.54 & 69.56  & 67.28 \\
    RQFormer~\cite{rqformer} & R50 & 41M & 67.31 & 55.23 & 74.19 & 82.74 & 44.49 & 78.56 & 39.85 & 70.27 & 79.84 & 75.10 & 80.38 & 45.64 & 58.51 & 88.91 & 68.10 & 75.73 & 85.52 & 57.17 & 53.54 & 65.05 & 67.31\\
    ReDiffDet~\cite{rediffdet} & ReR50 & - & 71.36 & 49.22 & 71.65 & \dunderline{87.88} & 47.12 & 79.28 & 33.35 & 73.37 & 83.74 & 70.29 & 80.38 & 43.63 & 57.17 & \textbf{89.52} & 72.39 & \textbf{79.81} & \textbf{89.03} & 57.34 & 57.32 & 67.23 & 68.05 \\
    GSDet~\cite{gsdet} & R50 & 106M & 81.07 & \dunderline{58.94} & 78.06 & 85.91 & \dunderline{48.64} & 77.55 & \textbf{42.51} & 73.25 & 86.18 & 75.04 & 78.98 & 41.29 & \dunderline{59.81} & 87.54 & \textbf{78.43} & 76.96 & 87.22 & 58.06 & \dunderline{57.30} & 68.25 & \dunderline{70.05} \\
    \midrule
    \textbf{O$^2$-RTDETR} & R34 & 31M & 71.58 & 50.71 & 79.10 & 83.75 & 48.09 & 77.85 & 37.19 & 74.22 & 84.34 & 75.62 & \dunderline{83.01} & 44.28 & 59.24 & 85.51 & 70.21 & 75.14 & 84.63 & \dunderline{62.03} & 54.74 & \dunderline{72.17} & 68.67 \\
    \textbf{O$^2$-RTDETR} & R50 & 42M & \textbf{79.36} & \textbf{59.61} & \dunderline{79.33} & \textbf{88.73} & \textbf{50.80} & \textbf{81.42} & \dunderline{42.34} & \textbf{79.64} & \textbf{89.53} & \dunderline{78.05} & \textbf{83.95} & \textbf{50.01} & \textbf{61.35} & 88.46 & 72.80 & \dunderline{77.34} & \dunderline{88.47} & 61.80 & \textbf{59.09} & \textbf{73.02} & \textbf{72.26} \\
    \textbf{O$^2$-DFINE-s} & HG-B0 & 10M & 66.89 & 47.97 & 77.73 & 83.51 & 43.64 & 78.37 & 32.82 & 69.65 & 84.01 & 77.17 & 82.04 & 45.14 & 56.78 & 85.91 & 73.74 & 71.63 & 82.82 & \textbf{62.18} & 52.62 & 66.22 & 67.05 \\
    \textbf{O$^2$-DFINE-m} & HG-B2 & 19M & \dunderline{76.83} & 51.90 & 78.22 & 85.52 & 46.93& \dunderline{80.79} & 37.72 & 77.23 & \dunderline{89.24} & 77.60 & 82.42 & \dunderline{49.78} & 58.00 & 87.16 & 67.20 & 77.27 & 86.55 & 61.01 & 54.65 & 71.25 & 69.86\\
    \bottomrule
    \end{tabular}
    }
    \label{DIOR-R-result}
    \end{table*}
}

\newcommand{\DOTAOneResults}{
\begin{table*}[!tb]
\caption{Performance comparisons on the \textbf{DOTA-v1.0} dataset. $^\clubsuit $ indicates multi-scale training and testing. FLOPs are computed for an input size of (1,3,1024,1024). FPS is measured on an NVIDIA RTX 2080Ti using TensorRT in FP16 mode with an input size of (1,3,1024,1024). The \textbf{best} and \dunderline{second-best} results are highlighted in \textbf{bold} and \dunderline{underlined}.}
\setlength{\tabcolsep}{1.2mm}
\centering
\resizebox{\linewidth}{!}{
\begin{tabular}{l|ccccc|ccccccccccccccc|c}
\toprule
\textbf{Methods} & \textbf{Back.} & \textbf{\#P} & \textbf{FLOPs} & \textbf{FPS} & \textbf{Pre.} & \textbf{PL} & \textbf{BD} & \textbf{BR} & \textbf{GTF} & \textbf{SV} & \textbf{LV} & \textbf{SH} & \textbf{TC} & \textbf{BC} & \textbf{ST} & \textbf{SBF} & \textbf{RA} & \textbf{HA} & \textbf{SP} & \textbf{HC} & \textbf{AP}$_\text{50}$ \\ 

\hline
\hline

\rowcolor{gray!15} \multicolumn{22}{l}{$\blacktriangledown$ \textit{Non-real-time Oriented Object Detectors}} \\
ARS-DETR~\cite{arsdetr} & R50 & 41M & 407G & - & IN & 86.97 &75.56 &48.32 &69.20 &77.92& 77.94 &87.69 &90.50& 77.31 &82.86 &60.28 &64.58 &74.88& 71.76& 66.62 &74.16\\
RO$^2$-DETR~\cite{ro2detr} & R50 & - & - & - & IN & 88.98 & 82.13 & 54.58 & 76.28 & 79.28 & 77.92 & 87.94 & 90.90 & 87.19 & 85.65 & 62.20 & 62.63 & 74.62 & 72.42 & 59.22 & 77.57\\
O-DETR~\cite{orienteddetr} & R50 & 57M & 563G & - & coco & 89.20 & \textbf{86.40} & \dunderline{57.70} &75.30 &81.10 &84.70 & \dunderline{89.10} &90.90 &86.10 &87.00& 59.50 &70.30 &79.30& 81.50& 68.80 &79.10\\
MessDet~\cite{messdet} & CSPnext & 13M & 191G & - & IN & 89.28 & 84.69 & 56.20 &66.18 & 81.37 & 85.33 &88.86 & 90.78 & 88.28& 87.62 &64.47& 65.59 &78.18& 82.10 &67.82 & 78.45\\
ReDiffDet~\cite{rediffdet} & ReR50 & - & - & - & IN & 86.32& 78.72& 51.56& 71.89& 80.05& 84.73& 88.75& 90.84& 85.92& 86.69 &59.48& 57.74 &75.19& 68.43 &65.37 & 75.45 \\
FRED~\cite{fred} & ReR50 & - & - & - & IN &  89.37& 82.12& 50.84& 73.89& 77.58& 77.38& 87.51& 90.82& 86.30 &84.25& 62.54& 65.10 &72.65 &69.55& 63.41 & 75.56 \\
GauCho$^\clubsuit $~\cite{gaucho} & R50 & 55M & 505G & - & IN & 88.96& 81.01& 57.39 &72.21 & \textbf{82.40} &85.41 &88.51 &90.85 &85.42 &86.40 &66.42 &70.19& 76.10 &80.42 &71.00 &78.85 \\
ACM$^\clubsuit $~\cite{acm}   & R50 & - & - & - & IN & \dunderline{89.84} & 85.50& 53.84& 74.78& 80.77& 82.81& 88.92& 90.82& 87.18& 86.53& 64.09& 66.27& 77.51& 79.62& 69.57& 78.53\\
GSDet~\cite{gsdet} & R50 & 106M & 433G & - & IN & 88.57 & 81.17 &51.70& 73.19& 80.22& 83.54& 88.64& 90.82& 84.10& 81.59& 59.48 &64.69& 75.08& 70.00 &63.33 &75.74 \\
O-RCNN~\cite{orientedrcnn} & R50 & 41M & 450G & - & IN & 89.46 & 82.12 & 54.78 & 70.86 & 78.93 & 83.00 & 88.20 & 90.90 & 87.50 & 84.68 & 63.97 & 67.69 & 74.94 & 68.84 & 52.28 & 75.87 \\
O-RCNN~\cite{orientedrcnn} & PKI~\cite{pkinet} &31M & 407G & - & IN &  89.72& 84.20& 55.81 &77.63 &80.25 &84.45& 88.12 &90.88& 87.57& 86.07& 66.86& 70.23 &77.47 &73.62 &62.94& 78.39 \\
O-RCNN~\cite{orientedrcnn} & LSK~\cite{lsknet} &31M & 375G & - & IN & 89.66& 85.52& \textbf{57.72} & 75.70 &74.95 &78.69 &88.24& 90.88 &86.79 &86.38& 66.92& 63.77& 77.77 &74.47& 64.82& 77.49\\

\hline
\hline

\rowcolor{gray!15} \multicolumn{22}{l}{$\blacktriangledown$ \textit{Non-end-to-end Real-time Oriented Object Detectors}} \\
YOLOX-s~\cite{yolox} & CSP-s & 9M & 68G & 237 & none & 86.43 & 82.94 & 47.92 & 65.51 & 78.02 & 84.97 & 89.07 & 90.84 & 87.99 & 86.01 & 58.29 & 63.63 & 73.23 & 69.81 & 59.51 & 74.94\\
YOLOX-m~\cite{yolox} & CSP-m & 25M & 188G & 154 & none & 86.84 & \dunderline{86.40} & 54.27 & 70.04 & 81.03 & \dunderline{85.68} & \textbf{89.27} & 90.71 & 86.32 & 86.17 & 60.31 & 65.52 & 75.66 & 72.68 & 59.53 & 76.70\\
PPYOLOE-s~\cite{ppyoloer} & CRN-s & 8M & 45G & 70 & IN & 88.80 & 79.24 & 45.92 & 66.88& 80.41& 82.95& 88.20& 90.61& 82.91& 86.37& 55.80& 64.11& 65.09 &79.50 & 50.43& 73.82\\
PPYOLOE-m~\cite{ppyoloer} & CRN-m & 23M & 131G & 55 & IN & 89.23& 79.92& 51.14& 72.94& 81.86& 84.56& 88.68& 90.85& 86.85& 87.48& 59.16& 68.34& 73.78& 81.72& 68.10&77.64\\
PPYOLOE-l~\cite{ppyoloer} & CRN-l & 52M & 291G & 48 & IN & 89.18& 81.00& 54.01& 70.22& 81.85& 85.16& 88.81& 90.81& 86.99& 88.01& 62.87& 67.87& 76.56& 79.13& 69.65&78.14\\
YOLOv5-n~\cite{yolov5} & CSP-n & 2M & 13G & 108 & coco & 89.82 & 75.73 & 44.75 & 62.78 & 79.91 & 82.80 & 87.91 & 90.83 & 79.15 & 87.12 & 50.54 & 63.11 & 67.49 & 80.07 & 56.87 & 73.26 \\ 
YOLOv5-s~\cite{yolov5} & CSP-s & 8M & 45G & 98 & coco & 89.26 & 84.53 & 51.36 &60.69 & 80.70 & 84.74 & 88.41 & 90.68 & 85.93& 87.59 & 59.62 & 65.18 & 74.53 & 81.71 & 66.91 & 76.79\\
YOLOv5-m~\cite{yolov5} & CSP-m & 22M & 129G & 72 & coco & 89.60 & 86.39 & 54.62 & 62.04 & 80.36& 85.20 & 88.40 & 90.80 & 80.54 & \dunderline{88.47} & 64.16 & 61.21 & 76.70 & \textbf{83.13} & 67.91 & 77.30\\
PSC-s~\cite{psc} & CSP-s & 8M & 45G & 98 & coco & 89.65 & 86.37 & 51.76 & 63.42 & 81.21 & 84.63 & 88.29 & 90.80 & 85.39 & 87.93 & 61.00 & 66.41 & 75.01 & 81.77 & 66.20 & 77.32\\
PSC-m~\cite{psc} & CSP-m & 22M & 129G & 72 & coco & \textbf{89.85} & 85.93 & 54.94 & 61.56 & \dunderline{81.89} & 85.47 & 88.37 & 90.73 & 86.90 & \textbf{88.79} & 63.90 & \textbf{68.92} & 76.82 & \dunderline{82.83} & 63.25 & 78.01\\
RTMDet-s~\cite{rtmdet} & CSP-s & 9M & 126G & 197 & IN & 89.18& 80.45& 52.09& 71.35& 81.55& 84.05& 88.79& 90.89& 87.83& 86.98& 59.58& 62.28& 75.90& 81.96& 61.04&76.93
\\
RTMDet-m~\cite{rtmdet} & CSP-m & 25M & 314G & 120 & IN & 89.17& 84.65& 53.92& 74.67& 81.48& 83.99& 88.71& 90.85& 87.43& 87.20& 59.39& 66.68& 77.71& 82.40& 65.28&78.24\\
RTMDet-l~\cite{rtmdet} & CSP-l & 52M & 612G & 86 & IN & 89.43& 84.21& 55.20& 75.06& 80.81& 84.53& 88.97& 90.90& 87.38& 87.25& 63.09& 67.87& 78.09& 80.78& 69.13&78.85\\
YOLOMS-xs~\cite{yoloms} & MS-xs & 4M & 73G & 139 & coco & 89.44&84.21&52.63&74.07&81.76&82.51&88.71&90.90&87.66&86.50&63.04&58.84&76.51&69.19&52.18 & 75.88 \\
YOLOMS-s~\cite{yoloms} & MS-s & 8M & 131G & 100 & coco &89.37 & 83.92 & 53.26 & 73.54 & 81.75 & 84.94 & 88.91 & 90.89 & \dunderline{88.34} & 87.86 & 61.99 & 56.97 & 77.54 & 75.65 & 53.26&76.55\\
\hline
\hline

\rowcolor{gray!15} \multicolumn{22}{l}{$\blacktriangledown$ \textit{End-to-end Real-time Oriented Object Detection}} \\ 
RTDETR-R & R18 & 20M & 147G & 233 & IN & 86.20 & 69.96 & 50.33 & 70.66 & 78.79 & 84.32 & 87.40 & 90.28 & 85.05 & 77.55 & 52.01 & 56.58 & 76.27 & 79.82 & 63.89 & 73.94\\
RTDETR-R & R34 & 31M & 228G & 174 & IN & 87.24 & 78.31 & 51.37 & 70.85 & 79.54 & 83.75 & 87.31 & 90.02 & 86.71 & 84.64 & 57.25 & 61.18 & 76.22 & 70.09 & 58.69 & 74.88\\
RTDETR-R & R50 & 42M & 339G & 119 & IN & 87.46 & 77.98 & 51.60 & 72.06 & 79.66 & 84.10 & 87.72 & 90.29 & 87.32 & 84.81 & 57.37 & 60.28 & 76.59 & 71.37 & 60.29 & 75.26\\
\rowcolor{gray!15} O$^2$-RTDETR & R18 & 20M & 147G & 233 & IN &88.96 & 84.23 & 51.98 & 71.53 & 80.19 & 84.98 & 88.40 & 90.87 & 87.86 & 85.47 & 57.69 & 66.27 & 77.19 & 74.62 & 69.42 & 77.31\\
\rowcolor{gray!15} O$^2$-RTDETR & R34 & 31M & 228G & 174 & IN &88.88 & 83.77 & 53.75 & 73.90 & 80.76 & 85.36 & 88.61 & 90.87 & 86.64 & 86.75 & 58.31 & 66.44 & 77.25 & 80.07 & 70.68  & 78.13\\
\rowcolor{gray!15} O$^2$-RTDETR & R50 & 42M & 339G & 119 & IN &89.27 & 84.06 & 56.50 & 75.85 & 80.02 & 84.06 & 88.05 & 90.85 & 85.32 & 86.68 & 63.19 & 64.96 & 77.76 & 77.80 & 72.38 & 78.45  \\
\rowcolor{gray!15} O$^2$-DEIM$^\clubsuit $ & R18 & 20M & 147G & 233 & coco & 88.60 & 83.38 & 54.00 & 79.62 & 80.13 & \textbf{85.93} & 88.57 & 90.90 & 88.11 & 85.77 & 67.55 & 66.67 & 77.11 & 77.99 & 77.96 & 79.49\\
\rowcolor{gray!15} O$^2$-DEIM$^\clubsuit $ & R34 & 31M & 228G & 174 & coco & 88.03 & 84.62 & 56.97 & \dunderline{80.02} & 81.05 & 84.74 & 88.62 & \dunderline{90.90} & \textbf{88.54} & 86.86 & \dunderline{68.80} & \dunderline{68.86} & \textbf{80.28} & 79.00 & 73.27 & \dunderline{80.04} \\
\rowcolor{gray!15} O$^2$-DEIM$^\clubsuit $ & R50 & 42M & 339G & 119 & coco & 88.14 & 84.89 & 57.13 & \textbf{80.34} & 81.21 & 84.54 & 88.67 & \textbf{90.90} & 88.31 & 86.97 & \textbf{69.83} & 68.80 & \dunderline{80.20} & 78.71 & \dunderline{73.56} & \textbf{80.15} \\
DFINE-R-s & HG-B0 & 10M & 60G & 297 & IN & 85.75 & 62.29 & 47.76 & 67.69 & 79.62 & 80.99 & 87.60 & 90.30 & 82.41 & 78.69 & 57.09 & 55.67 & 74.43 & 72.62 & 63.63 & 72.44 \\
DFINE-R-m & HG-B2 & 19M & 142G & 176 & IN & 86.33 & 69.91 & 49.68 & 66.18 & 79.69 & 84.47 & 88.28 & 90.85 & 87.80 & 78.41 & 55.04 & 52.89 & 76.32 & 71.66 & 71.31 & 73.92 \\
DFINE-R-l & HG-B4 & 31M & 229G & 132 & IN & 85.72 & 72.69 & 50.50 & 66.14 & 79.76 & 84.55 & 88.30 & 90.90 & 86.52 & 77.60 & 55.50 & 55.61 & 75.89 & 71.95 & 69.78 & 74.09 \\
\rowcolor{gray!15} O$^2$-DFINE-s & HG-B0 & 10M & 60G & 297 & IN & 88.84 & 75.59 & 53.97 & 69.09 & 80.02 & 85.02 & 88.51 & 90.85 & 87.71 & 87.63 & 56.32 & 63.23 & 77.10 & 69.23 & 68.88 & 76.14 \\
\rowcolor{gray!15} O$^2$-DFINE-m & HG-B2 & 19M & 142G & 176 & IN & 89.42 & 81.80 & 53.92 & 72.89 & 79.42 & 83.90 & 88.21 & 90.88 & 86.56 & 86.42 & 58.51 & 66.03 & 77.41 & 78.28 & 70.46 & 77.61 \\
\rowcolor{gray!15} O$^2$-DFINE-l & HG-B4 & 31M & 229G & 132 & IN & 89.08 & 80.18 & 55.17 & 69.73 & 80.53 & 83.77 & 88.46 & 90.86 & 87.55 & 86.86 & 60.14 & 68.68 & 77.17 & 77.79 & 70.03 & 77.73 \\
\bottomrule
\end{tabular}}
\label{DOTA-1.0-result}
\end{table*}
}
\newcommand{\TableDOTAVonefiveResult}{
    \begin{table*}[!t]
    \setlength{\tabcolsep}{5pt}
    \caption{Performance comparisons on the \textbf{DOTA-v1.5} dataset. The \textbf{best} and \dunderline{second-best} results are highlighted in \textbf{bold} and \dunderline{underlined}. Results are reported under single-scale training and testing.}
    \resizebox{\linewidth}{!}
    {
        \begin{tabular}{lccccccccccccccccccc}
        \toprule
        \textbf{Method} & \textbf{Back.} & \textbf{\#P} & \textbf{PL} & \textbf{BD} & \textbf{BR} & \textbf{GTF} & \textbf{SV} & \textbf{LV} & \textbf{SH} & \textbf{TC} & \textbf{BC} & \textbf{ST} & \textbf{SBF} & \textbf{RA} & \textbf{HA} & \textbf{SP} & \textbf{HC} & \textbf{CC} & \textbf{AP$_{50}$} \\
        \midrule
        GauCho~\cite{gaucho} & R50 & 55M &76.42 & 72.78 & 48.42 & 59.72 & 61.65 & 75.19 & 84.83 & \dunderline{90.88} & 76.44 & 73.88 & 56.75 & 69.51 & 62.98 & 67.79 & 50.55 & 13.65 & 65.09 \\
        ReDet~\cite{redet} & ReR50 & 32M & 79.20 & 82.81 & 51.92 & 71.41 & 52.38 & 75.73 & 80.92 & 90.83 & 75.81 & 68.64 & 49.29 & 72.03 & 73.36 & 70.55 & 63.33 & 11.53 & 66.86 \\
        DCFL~\cite{dcfl} & R50 & 36M & - & - & - & - & 56.72 & - & 80.87 & - & - & 75.65 & - & - & - & - & - & - & 67.37 \\
        RQFormer~\cite{rqformer} & R50 & 41M & 78.44 & 75.43 & 48.84 & 63.62 & 62.16 & 78.41 & 88.94 & 90.84 & 75.29 & 81.83 & 57.05 & 66.44 & 73.95 & 72.69 & 51.64 & 13.37 & 67.43\\
        GSDet~\cite{gsdet} & R50 & 106M & \dunderline{80.48} & 72.51 & 49.11 & 71.19 & 57.75 & 78.96 & 88.93 & 90.85 & 79.44 & 77.52 & 55.77 & 65.56 & 71.36 & 70.40 & 65.41 & 10.26 & 67.84 \\
        FRED~\cite{fred} & R50 & - & 79.60 & 81.44 & 52.60 & 72.57 & 58.07 & 74.82 & 86.12 & 90.81 & 82.13 & 74.84 & 53.37 & 72.93 & 69.51 & 69.91 & 54.82 & 19.27 & 68.30\\
        O-RCNN~\cite{orientedrcnn} & LSK-s~\cite{lsknet} & 31M & 72.05 & 84.94 & \dunderline{55.41} & \textbf{74.93} & 52.42 & 77.45 & 81.17 & 90.85 & 79.44 & 69.00 & 62.10 & \dunderline{73.72} & 77.49 & 75.29 & 55.81 & 42.19 & 70.26 \\
        O-RCNN~\cite{orientedrcnn} & PKI-s~\cite{pkinet} & 31M & 80.31 & \dunderline{85.00} & \textbf{55.61} & \dunderline{74.38} & 52.41 & 76.85 & 88.38 & 90.87 & 79.04 & 68.78 & \textbf{67.47} & 72.45 & 76.24 & 74.53 & 64.07 & 37.13 & 71.47 \\
        RHINO~\cite{rhino} & R50 & 48M & 77.96 & 83.22 & 55.30 & 72.14 & 65.07 & 78.95 & 89.22 & 90.78 & 80.90 & \textbf{83.48} & 61.58 & \textbf{74.17} & 77.21 & 71.99 & 60.16 & 29.26 & 71.96 \\

        MessDet~\cite{messdet} & CSPnext & 13M & \textbf{80.61} & 84.99 & 54.56 & 65.36 & \textbf{73.60} & \textbf{82.69} & \textbf{89.73} & 90.83 & 82.46 & 82.03 & 59.35 & 72.96 & \dunderline{77.66} & 75.14 & 65.09 & 21.12 & 72.38\\
        RTMDet-s~\cite{rtmdet} & CSPnext-s & 9M & 80.05 & 84.36 & 50.65 & 72.04 & 59.54 & 81.79 & 89.22 & \textbf{90.90} & \dunderline{83.07} & 76.27 & 56.82 & 72.13 & 76.25 & \textbf{77.04} & 65.66 & 32.84 & 71.79 \\
        RTMDet-m~\cite{rtmdet} & CSPnext-m & 25M & 80.34 & \textbf{86.00} & 54.02 & 72.98 & 63.21 & \dunderline{82.09} & \dunderline{89.46} & 90.87 & \textbf{85.12} & 76.69 & \dunderline{63.12} & 72.14 & \textbf{77.91} & \dunderline{76.04} & 71.57 & 32.24 & \dunderline{73.36} \\
        \midrule
        \textbf{O$^2$-RTDETR} & R34 & 31M &73.92 & 84.31 & 54.07 & 69.62 & \dunderline{65.92} & 80.48 & 88.91 & 90.82 & 79.37 & 80.37 & 55.95 & 69.60 & 74.91 & 73.70 & 74.38 & \dunderline{34.20} & 71.91\\
        \textbf{O$^2$-RTDETR} & R50 & 42M & 80.43 & 83.97 & 53.37 & 72.14 & 66.58 & 80.10 & 89.40 & 90.80 & 78.94 & \dunderline{83.45} & 58.44 & 70.38 & 76.73 & 75.56 & \dunderline{75.11} & \textbf{44.77} & \textbf{73.76}\\
        \textbf{O$^2$-DFINE-s} & HG-B0 & 10M & 74.73 & 81.99 & 53.87 & 66.56 & 64.88 & 77.82 & 88.39 & 90.74 & 77.57 & 78.65 & 55.70 & 68.69 & 69.33 & 70.79 & 71.07 & 32.83 & 70.22 \\
        \textbf{O$^2$-DFINE-m} & HG-B2 & 19M & 77.64 & 84.11 & 54.08 & 69.34 & 65.58 & 80.77 & 88.64 & 90.60 & 79.13 & 80.64 & 57.64 & 69.86 & 74.56 & 73.54 & \textbf{76.19} & 32.10 & 72.15\\
        \bottomrule
        \end{tabular}
    }

    \label{DOTA-1.5-result}
    \end{table*}
}
\newcommand{\TableFAIRResult}{
    \begin{table*}[!htbp]
    \caption{Performance comparisons on the \textbf{FAIR1M-v1.0} dataset. The \textbf{best} and \dunderline{second-best} results are highlighted in \textbf{bold} and \dunderline{underlined}. Results are reported under single-scale training and testing.}
        \resizebox{\linewidth}{!}
        {
            \begin{tabular}{lcccccccccccccccccccc}
            \toprule
            \textbf{Method} & \textbf{Back.} & \textbf{\#P} & \textbf{B737} & \textbf{B747} & \textbf{B777} & \textbf{B787} & \textbf{C919} & \textbf{A220} & \textbf{A321} & \textbf{A330} & \textbf{A350} & \textbf{ARJ21} & \textbf{PS} & \textbf{MB} & \textbf{FB} & \textbf{TB} & \textbf{ES} & \textbf{LCS} & \textbf{DCS} &   \\
            \midrule
            RetinaNet~\cite{retinanet} & R101 & 36M & 38.46 & 55.36 & 24.75 & 51.81 & 0.81 & 40.50 & 41.06 & 18.02 & 19.94 & 1.70 & 9.57 & 22.55 & 1.33 & 16.37 & 19.11 & 14.26 & 24.70\\
            CasRCNN~\cite{cascadercnn} & R101 & 69M & 40.42 & 52.86 & \dunderline{29.07} & 52.47 & 0.00 & 44.37 & 38.35 & 26.55 & 17.54 & 0.00 & 12.10 & 28.84 & 0.71 & 15.35 & 18.53 & 14.63 & 25.15\\
            FasterRCNN~\cite{faster_rcnn} & R101 & 41M & 36.43 & 50.68 & 22.50 & 51.86 & 0.01 & 47.81 & 43.83 & 17.66 & 19.95 & 0.13 & 9.81 & 28.78 & 1.77 & 17.65 & 16.47 & 16.19 & 27.06\\
            CHODNet~\cite{fair1m} & R101 & - & 40.23 & 53.39 & \textbf{29.65} & 54.04 & 0.02 & 46.26 & 43.12 & 27.61 & 20.89 & 0.01 & 12.34 & 29.34 & 1.71 & 17.77 & 17.72 & 16.78 & 27.51\\
            RoITrans~\cite{roi_transformer} & R101 & 55M & 39.58 & 73.56 & 18.32 & \dunderline{56.43} & 0.00 & 47.67 & 49.91 & 27.64 & 31.79 & 0.00 & 14.31 & 28.07 & 1.03 & 14.32 & 15.97 & 18.04 & 26.02 \\
            O-RCNN~\cite{orientedrcnn} & R50 & 41M & 35.17 & 85.17 & 14.57 & 47.68 & 11.68 & 46.55 & \dunderline{68.18} & 68.60 & 70.21 & 25.32 & 13.77 & 60.42 & 9.10 & \dunderline{36.83} & 11.32 & \dunderline{21.86} & 38.22\\
            \textbf{O$^2$-RTDETR} & R34 & 31M & \dunderline{41.03} & 85.87 & 17.00 & 49.27 & 23.96 & 49.74 & 63.83 & 67.11 & 70.60 & \textbf{37.85} & \dunderline{14.78} & 62.73 & 9.73 & 30.09 & 13.28 & 21.69 & 39.50   &  \\
            \textbf{O$^2$-RTDETR} & R50 & 42M &39.42 & \textbf{86.87} & 20.94 & \textbf{57.01} & \dunderline{26.62} & 47.68 & 67.39 & 67.73 & \textbf{78.37} & \dunderline{34.06} & \textbf{15.90} & \textbf{65.19} & \dunderline{11.60} & 34.06 & \textbf{19.54} & \textbf{25.25} & \textbf{42.40} \\
            \textbf{O$^2$-DFINE-s} & HG-B0 & 10M & 39.54 & 84.20 & 18.60 & 50.56 & 25.15 & \dunderline{50.17} &64.30 & \dunderline{68.89} & 71.81 & 33.49 &12.73& 60.65 & 9.21 & 29.80 & 11.20 & 16.64 & 35.98 \\
            \textbf{O$^2$-DFINE-m} & HG-B2 & 19M & \textbf{41.56} & \dunderline{86.38} & 14.81 & 53.31 & \textbf{28.37} & \textbf{53.27} & \textbf{71.28} & \textbf{69.75} & \dunderline{77.43} & 27.52 & 11.58 & \dunderline{64.18} & \textbf{12.21} & \textbf{38.86} & \dunderline{19.44} & 20.26 & \dunderline{40.59} \\
            \midrule
            \textbf{Method} & \textbf{Back.} & \textbf{\#P} & \textbf{WS}   & \textbf{SC}   & \textbf{Bus}  & \textbf{CT}   & \textbf{DT}   & \textbf{VAN}  & \textbf{TRI}  & \textbf{TRC}  & \textbf{EX}   & \textbf{TT}    & \textbf{BC} & \textbf{TC} & \textbf{FF} & \textbf{BF} & \textbf{IS} & \textbf{RA}  & \textbf{BR}  & \textbf{AP$_{50}$} \\
            \midrule
            RetinaNet~\cite{retinanet} & R101 & 36M & 15.37 & 65.20 & 22.42 & 44.17 & 35.37 & 52.44 & \textbf{19.17} & 1.28 & 17.03 & 28.98 & 50.58 & 81.09 & 52.50 & 66.76 & \textbf{60.13} & 17.41  & 12.58 & 30.67 \\
            CasRCNN~\cite{cascadercnn} & R101 & 69M & 14.53 & 68.19 & 28.25 & 48.62 & 40.40 & 58.00 & 13.66 & 0.91 & 16.45 & \dunderline{30.27} & 38.81 & 80.29 & 48.21 & 67.90 & 55.67 & 20.35  & 12.62 & 31.18 \\
            FasterRCNN~\cite{faster_rcnn}& R101 & 41M & 13.16 & 68.42 & 28.37 & \dunderline{51.24} & 43.60 & 57.51 & 15.03 & 3.04 & \dunderline{17.99} & 29.36 & \textbf{58.26} & \textbf{82.67} & 54.50 & 71.71 & 59.86 & 16.92 & 11.87 & 32.12 \\
            CHODNet~\cite{fair1m} & R101 & - & 13.68 & \textbf{70.12} & 28.38 & 49.11 & 44.02 & 60.78 & 14.48 & 4.96 & 17.60 & 30.09 & 47.91 & 82.11 & 54.10 & 69.97 & \dunderline{59.91} & 19.57 & 14.38 & 32.93\\
            RoITrans~\cite{roi_transformer} & R101 & 55M & 12.97 & 68.80 & \dunderline{37.41} & \textbf{53.96} & \dunderline{45.68} & 58.39 & \dunderline{16.22} & \dunderline{5.13} & \textbf{22.17} & \textbf{46.71} & \dunderline{54.84} & 80.35 & 56.68 & 69.07 & 58.44 & 18.58 & 31.81 & 35.29\\
            O-RCNN~\cite{orientedrcnn} & R50 & 41M & 22.67 & 57.62 & 24.40 & 40.84 & 45.20 & 54.01 & 15.46 & 2.37 & 13.55 & 0.24 & 48.18 & 78.45 & \dunderline{60.79} & \dunderline{88.43} & 57.90 & 17.57 & 28.63 & 38.85\\
             \textbf{O$^2$-RTDETR} & R34 & 31M & 37.26 & 66.56 & 27.34 & 41.78 & 42.60 & 63.13 & 7.67 & 1.88 & 16.10 & 1.16 & 44.16 & 80.83 & 57.66 & 87.07 & 52.52 & \dunderline{20.59} & 28.83 & 40.45\\
             \textbf{O$^2$-RTDETR} & R50 & 42M & \textbf{39.51} & \dunderline{68.81} & \textbf{41.23} & 46.21 & \textbf{48.86} & \textbf{66.31} & 11.14 & 3.64 & 16.31 & 2.47 & 47.51 & 80.36 & 59.08 & \textbf{88.49} & 55.19 & 18.30 & \textbf{33.27} & \textbf{43.14}\\
             \textbf{O$^2$-DFINE-s} & HG-B0 & 10M & 34.52 & 63.41 & 24.38 & 38.61 & 41.19 & 60.55 & 7.95 & 4.71 & 15.27 & 0.51 & 44.04 & 82.08 & 58.34 & 87.30 & 52.43 & \textbf{22.18} & 31.89 & 39.77 \\
             \textbf{O$^2$-DFINE-m} & HG-B2 & 19M & \dunderline{39.13} & 66.37 & 24.86 & 41.82 & 44.08 & \dunderline{63.31} & 8.67 & \textbf{8.61} & 12.38 & 0.95 & 48.74 & \dunderline{82.63} & \textbf{63.21} & 86.98 & 54.74 & 18.97 & \dunderline{32.24} & \dunderline{42.01} \\
             \bottomrule
            \end{tabular}
    }
    \label{FAIR1M-1.0-result}
    \end{table*}
}
\newcommand{\EachProposedModuleOnDOTA}{
    \begin{table}[!t]
        \centering
        \caption{\textbf{Step-by-step improvements} from the baseline model to O$^2$-DFINE-s, O$^2$-RTDETR-R18, and O$^2$-DEIM-R18 on \textbf{DOTA-v1.0.}}
        \begin{tabular}{l|cccc}
            \toprule
            \textbf{O$^2$-DFINE-s}  & \textbf{AP$_{50}$}  & \textbf{\#P} & \textbf{FPS} & \textbf{FLOPs} \\
            \midrule
            DFINE-R-s & 72.44 & 10M & 297 & 60G \\
            + oriented contrastive denoising & 73.72 & 10M & 297 & 60G\\
            + angle distribution refinement & 75.05 & 10M & 297 & 60G\\
            + Chamfer distance cost & 76.14 & 10M & 297 & 60G\\
            \hline \midrule
            \textbf{O$^2$-RTDETR-R18}  & \textbf{AP$_{50}$}  & \textbf{\#P} & \textbf{FPS} & \textbf{FLOPs} \\
            \midrule
            RTDETR-R-R18 & 73.94 & 20M & 233 & 147G \\
            + oriented contrastive denoising & 75.72 & 20M & 233 & 147G \\
            + Chamfer distance cost & 77.31 & 20M & 233 & 147G \\
            \hline \midrule
            \textbf{O$^2$-DEIM-R18}  & \textbf{AP$_{50}$}  & \textbf{\#P} & \textbf{FPS} & \textbf{FLOPs} \\
            \midrule
            O$^2$-RTDETR-R18 & 77.31 & 20M & 233 & 147G \\
            + Matchability-Aware Loss and \\
            \quad Multi-scale training  & 79.49 & 20M & 233 & 147G \\
            \bottomrule
        \end{tabular}
      \label{EachProposedModuleOnDOTA}
      \end{table}
}

\newcommand{\AblationCost}{
    \begin{table}[!htbp]
        \centering
        \caption{Ablation study of the \textbf{Chamfer distance cost} for bipartite matching in \textbf{O$^2$-DFINE-s} on \textbf{DOTA-v1.0}.}
    \resizebox{\linewidth}{!}{
        \begin{tabular}{ccccc|c}
        \toprule
        \textbf{Distance} & \textbf{Cost}   & \textbf{Distance} & \textbf{Loss}   & \textbf{Point}  & \multirow{2}{*}{\textbf{AP$_{50}$}} \\
        \textbf{Cost}     & \textbf{Weight} & \textbf{Loss}     & \textbf{Weight} & \textbf{Number} &                     \\
        \midrule
        KL Divergence &  2.0  & KLD & 2.0  & None  & 73.69\\
        L1 & 5.0 &  L1 & 5.0 & None  & 74.28 \\
        Hausdorff & 5.0  & L1 &   5.0   & 4 & 75.05\\
        \textbf{Chamfer} & \textbf{5.0}  & \textbf{L1} & \textbf{5.0} & \textbf{4} & \textbf{76.14} \\
        \hline \midrule
        Chamfer & 3.0  & L1 &   3.0   & 4 & 75.18  \\
        Chamfer & 4.0  & L1 &   4.0   & 4 & 75.40  \\
        \textbf{Chamfer} & \textbf{5.0}  & \textbf{L1} & \textbf{5.0} & \textbf{4} & \textbf{76.14} \\
        Chamfer & 6.0  & L1 &   6.0   & 4 &  75.69 \\
        \hline \midrule
        \textbf{Chamfer} & \textbf{5.0}  & \textbf{L1} & \textbf{5.0} & \textbf{4} & \textbf{76.14} \\
        Chamfer & 5.0  & L1 &   5.0   & 16 & 75.80 \\
        Chamfer & 5.0  & L1 &   5.0   & 32 & 75.54 \\
        \bottomrule
        \end{tabular}
    }
        \label{ablation_cost}
      \end{table}
}

\newcommand{\AblationDenoise}{
    \begin{table}[!t]
        \centering
        \caption{Ablation study of the \textbf{oriented contrastive denoising} in \textbf{O$^2$-DFINE-s} on \textbf{DOTA-v1.0}.}
          \begin{tabular}{c|cccc}
          \toprule
          \multirow{3}{*}{\textbf{Box noise}}    & $\lambda_1 $   & 0.5    &  \textbf{1.0}   &  2.0   \\
                                                 & $\lambda_2 $   & 1.0    &  \textbf{2.0}   &  4.0   \\
                                                 & AP$_{50}$      & 75.33  &  \textbf{76.14} &  75.12 \\
          \hline \midrule
          \multirow{3}{*}{\textbf{Angle noise}}  & $\lambda_3 $   & 3.0    & 5.0   & 9.0   \\
                                                 & $\lambda_4 $   & 6.0    & 10.0   & 18.0   \\
                                                 & AP$_{50}$      & 74.29  & 74.53 & 74.71  \\          
          \hline \midrule
          \multirow{5}{*}{\textbf{Geometric noise}}   & $\lambda_1 $  & 0.5   & 1.0   & 2.0   \\
                                                      & $\lambda_2 $  & 1.0   & 2.0   & 4.0   \\
                                                      & $\lambda_3 $  & 9.0   & 9.0   & 9.0   \\
                                                      & $\lambda_4 $  & 18.0   & 18.0   & 18.0   \\
                                                      & AP$_{50}$     & 75.05 & 75.72 & 74.88 \\  
          \hline \midrule
          \multirow{3}{*}{\textbf{Probability noise}}    & $\lambda_5 $  & 0.2 & 0.3 & 0.4 \\
                                                         & $\lambda_6 $  & 0.4 & 0.6 & 0.8 \\
                                                         & AP$_{50}$     &  74.32  &  74.56  & 74.49  \\  
          \hline \midrule
          \multirow{2}{*}{\textbf{Box noise}}     & Denoising query     &  100   &  \textbf{200}    & 300  \\
                                                  & AP$_{50}$           &  75.83 &  \textbf{76.14}  & 76.10 \\  
          \bottomrule
          \end{tabular}
            \label{ablation_denoise}
      \end{table}
}

\newcommand{\AblationAngleDistriRefine}{
    \begin{table}[!t]
        \centering
        \caption{Ablation study of \textbf{angle distribution refinement} in \textbf{O$^2$-DFINE-s} on \textbf{DOTA-v1.0}.}
          \begin{tabular}{c|ccccc}
          \toprule
          \textbf{Number of bins} $N$ & 8   & 16   & \textbf{32}   & 64 & 128   \\
          \midrule
          AP$_{50}$ & 75.78 & 75.86 & \textbf{76.14} & 75.98 & 75.91 \\
          \hline \midrule
          $a,c$ \textbf{of} $\mathcal{A}(\cdot)$ \textbf{in}~\eqref{weight_fun} & $\frac{1}{4},\frac{1}{4}$ & $\frac{1}{2},10$ & $\mathbf{\frac{1}{2}},\mathbf{\frac{1}{4}}$ & $\frac{1}{2},\frac{1}{8}$ & $1,\frac{1}{4}$ \\
          \midrule
          AP$_{50}$ & 75.27 & 75.64 & \textbf{76.14} & 75.98 & 75.92 \\
          \bottomrule
          \end{tabular}
            \label{ablation_angle_distri_refine}
      \end{table}
}

\usepackage{graphicx}

\newcommand{\latency}{
    \begin{figure}[!t]
    \centering
    \includegraphics[width=0.95\linewidth]{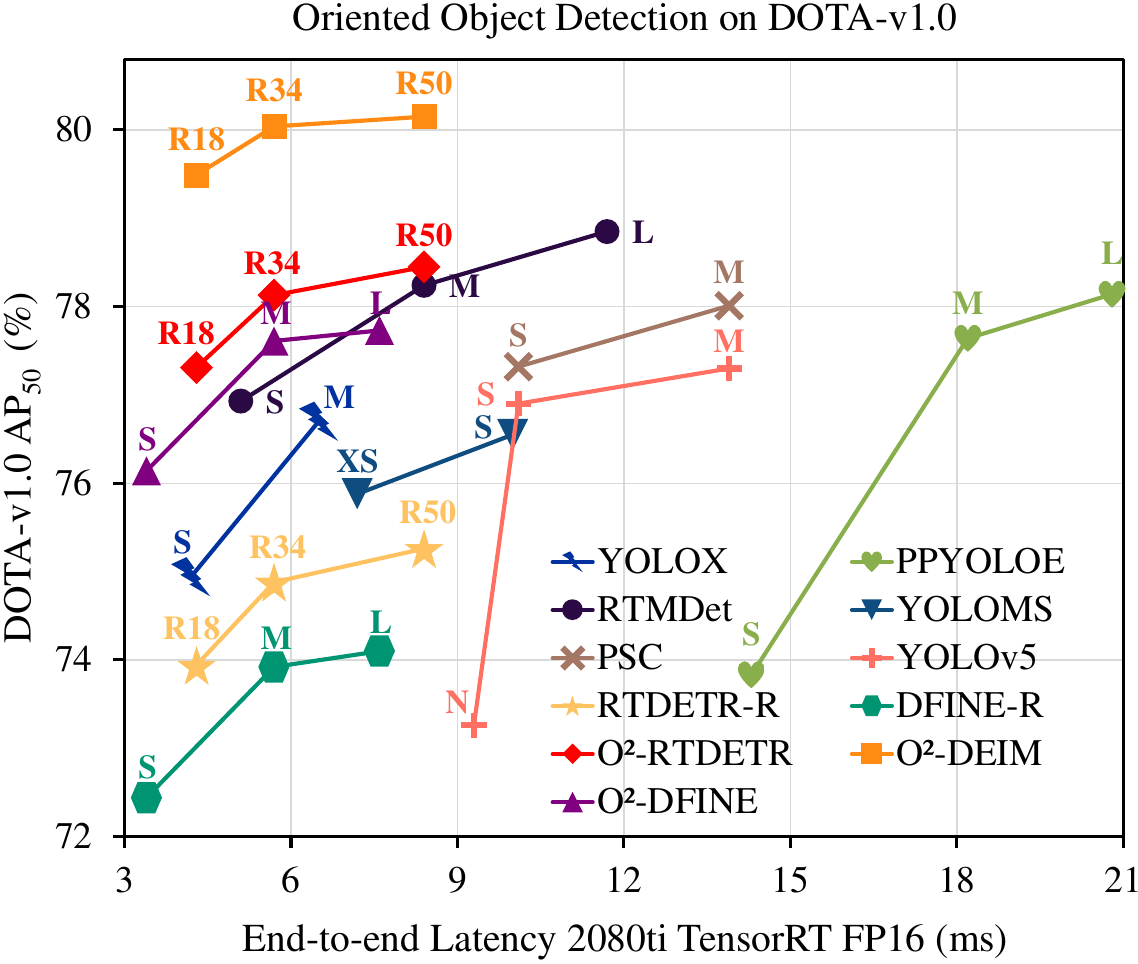}
    \caption{Compared to existing real-time object detectors, the family of O$^2$-DFINE, O$^2$-RTDETR, and O$^2$-DEIM achieves competitive performance.}
    \label{latency}
    \end{figure}

}

\newcommand{\differentConfThre}{
    \begin{figure}[!t]
    \centering
    \includegraphics[width=0.95\linewidth]{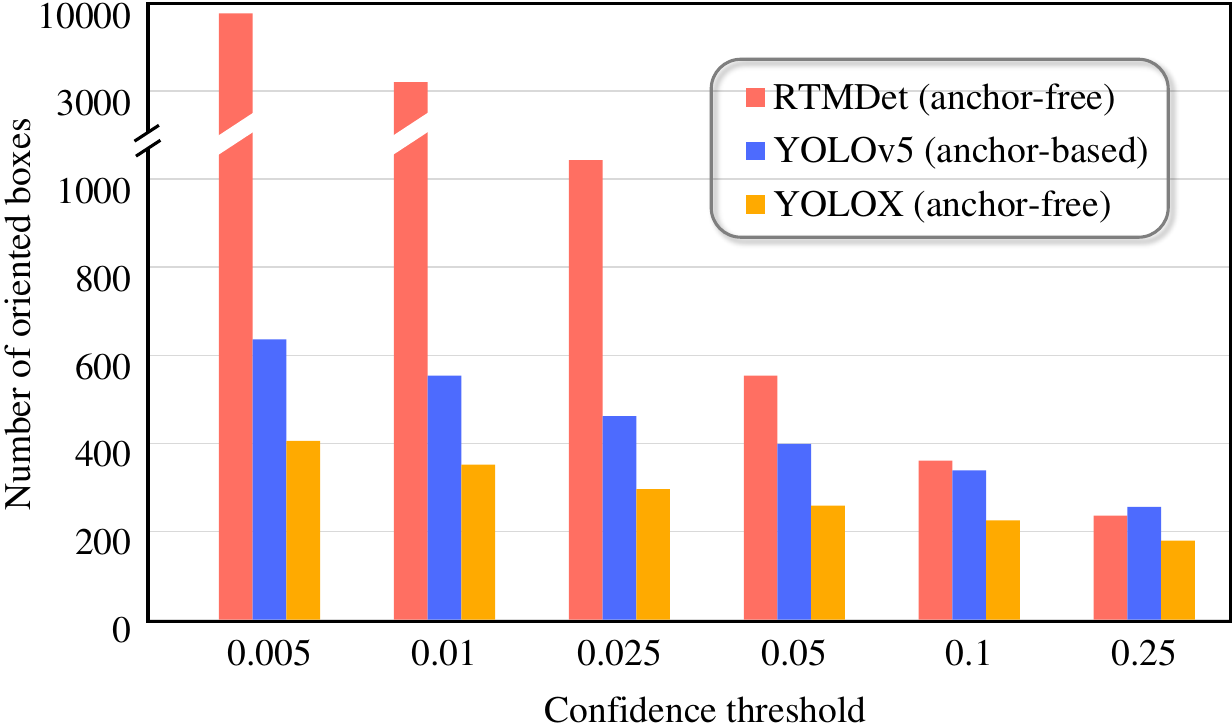}
    \caption{The number of remaining oriented boxes above different confidence thresholds in rotated NMS.}
    \label{differentConfThre}
    \vspace{-0.5em}
    \end{figure}
}

\newcommand{\NMSExeTime}{
    \begin{figure}[!t]
    \centering
    \includegraphics[width=0.95\linewidth]{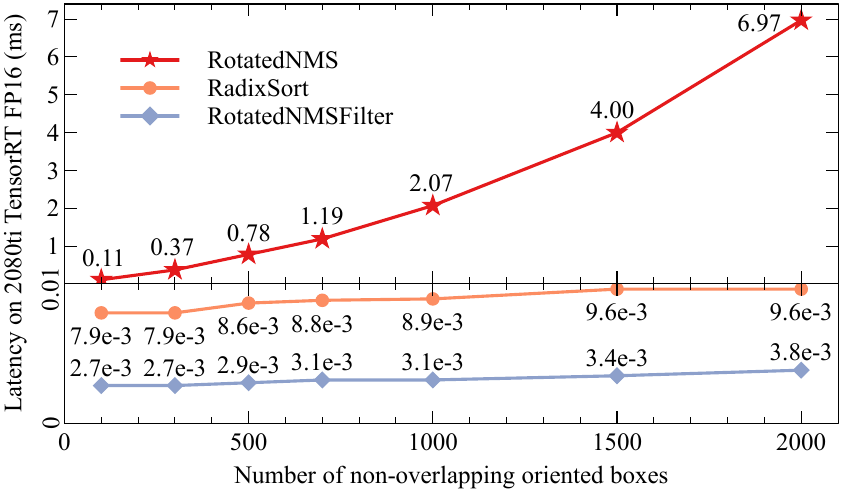}
    \caption{Execution time of rotated NMS with different numbers of non-overlapping oriented boxes on the 2080ti (TensorRT FP16).}
    \label{NMSExeTime}
    \vspace{-0.5em}
    \end{figure}
}

\newcommand{\angleDistribution}{
    \begin{figure*}[!t]
    \centering
    \includegraphics[width=\linewidth]{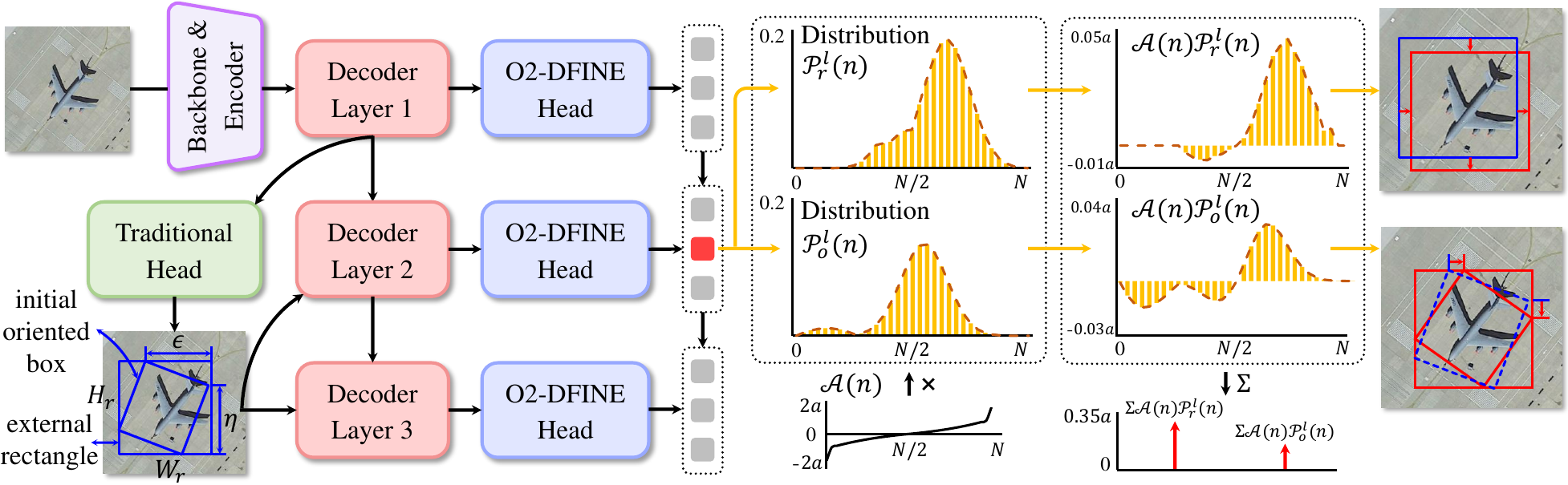}
    \caption{Overview of O$^2$-DFINE with Angle Distribution Refinement. Oriented boxes are learned in a decoupled manner using probability distributions as fine-grained representations of the external rectangle and vertex offsets, which are iteratively refined across decoder layers in a residual fashion.}
    \label{angleDistribution}
    \vspace{-1em}
    \end{figure*}
}

\newcommand{\charmferDistance}{
    \begin{figure*}[!t]
    \centering
    \includegraphics[width=0.95\linewidth]{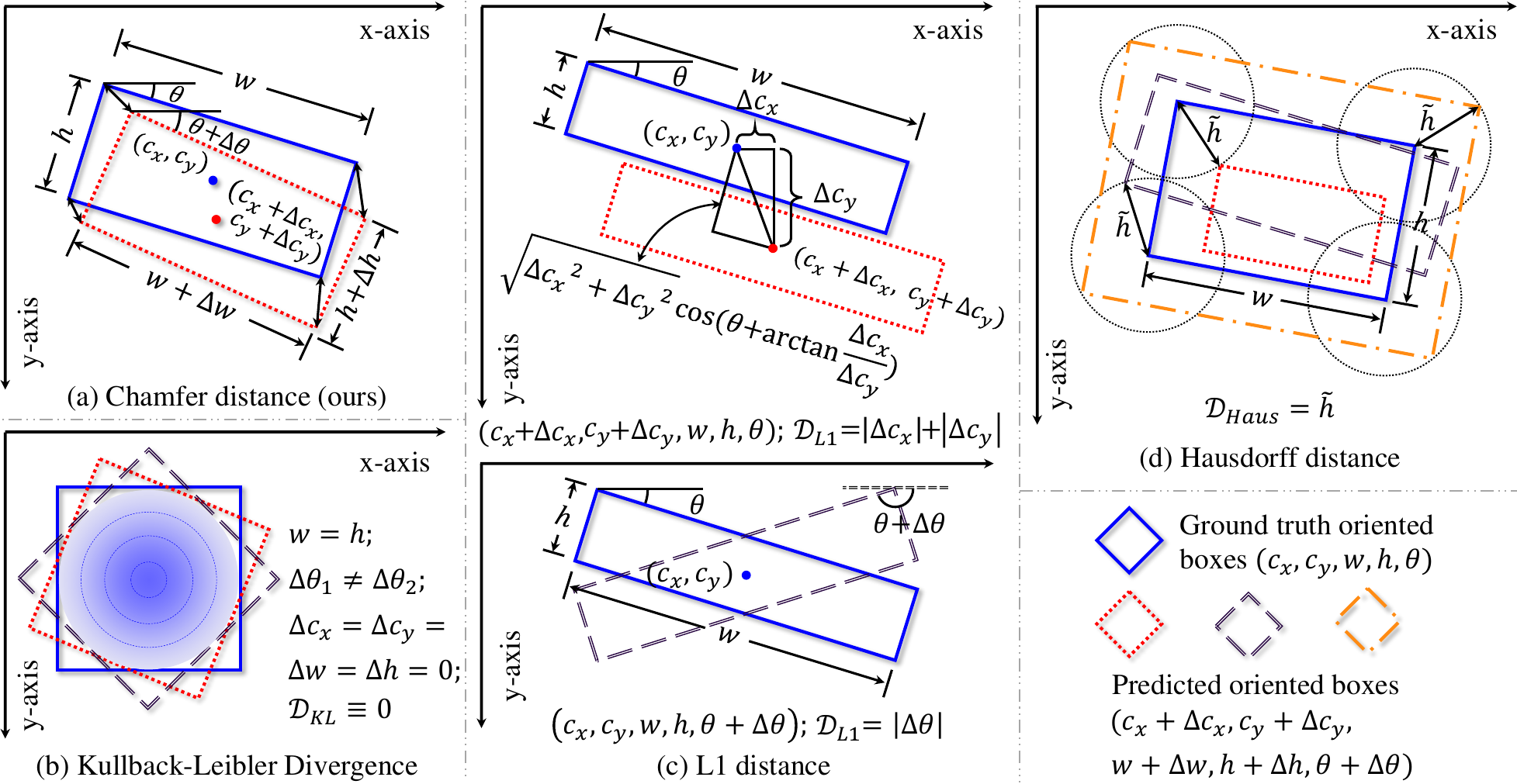}
    \caption{Distance cost for bipartite matching. (a) Chamfer distance (ours). (b) Kullback-Leibler Divergence. (c)  L1 distance. (d) Hausdorff distance.}
    \label{charmferDistance}
    \vspace{-0.5em}
    \end{figure*}
}

\newcommand{\contrastiveDenoise}{
    \begin{figure*}[!t]
    \centering
    \includegraphics[width=\linewidth]{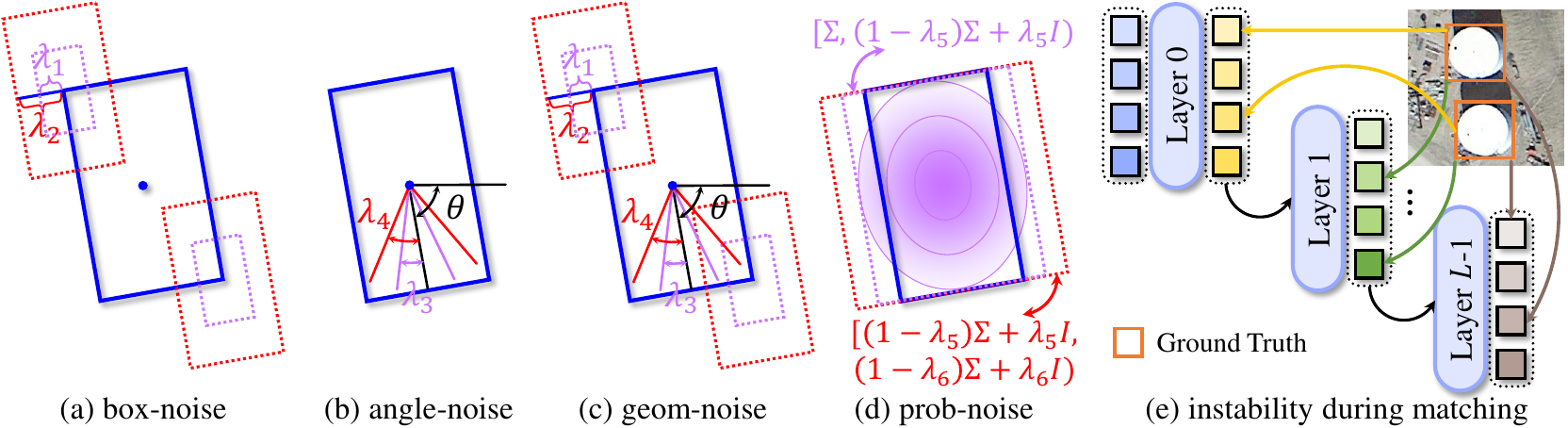}
    \caption{Oriented Contrastive Denoising. (a) Box noise. (b) Angle noise. (c) Geometric noise. (d) Probability noise. (e) Instability during mathcing.}
    \label{contrastiveDenoise}
    \vspace{-0.5em}
    \end{figure*}
}

\newcommand{\instability}{
    \begin{figure}[!t]
    \centering
    \includegraphics[width=0.95\linewidth]{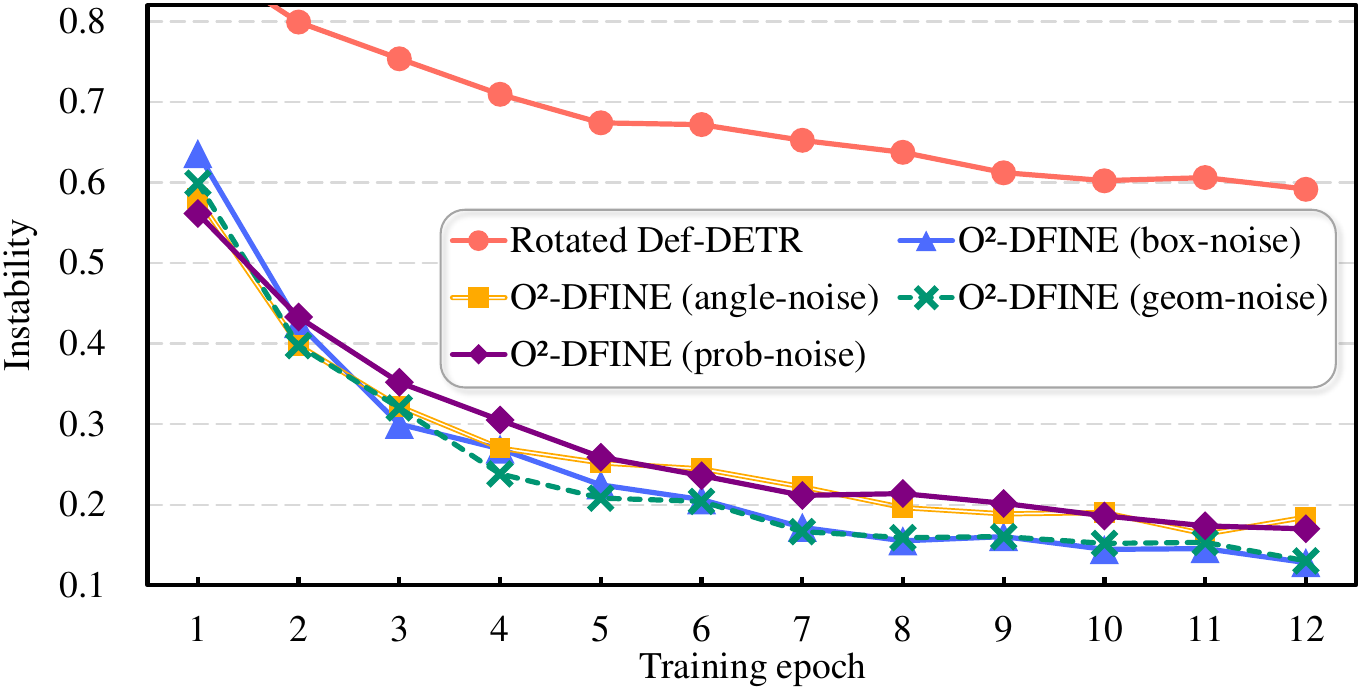}
    \caption{The instability metric of Oriented Contrastive Denoising.}
    \label{instability}
    \end{figure}
}

\definecolor{mypurple}{RGB}{202, 115, 255}

\newcommand{\visualContrastiveDenoise}{
    \begin{figure}[t]
      \centering
       \includegraphics[width=0.95\linewidth]{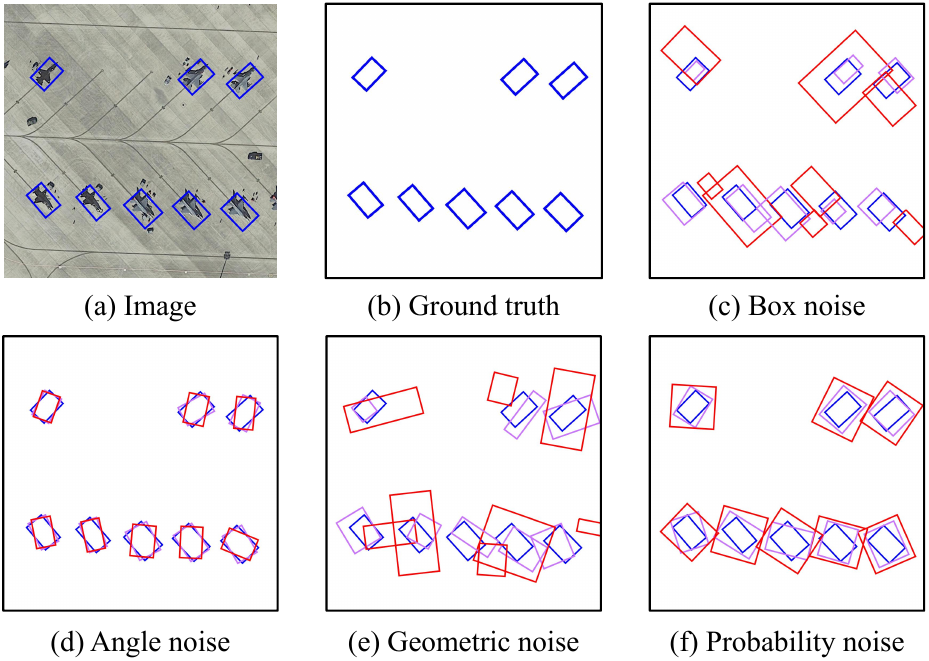}
        \parbox{0.75\linewidth}{\footnotesize
          \raisebox{-1ex}{\rotatebox{45}{\tikz{\draw[blue, thick] (0,0) rectangle (2ex,2ex);}}} ground truth \hfill
          \raisebox{-1ex}{\rotatebox{45}{\tikz{\draw[mypurple, thick] (0,0) rectangle (2ex,2ex);}}} positive \hfill
          \raisebox{-1ex}{\rotatebox{45}{\tikz{\draw[red, thick] (0,0) rectangle (2ex,2ex);}}} negtive
        }
       \caption{Qualitative analysis of Oriented Contrastive Denoising. (a) An image example. (b) Ground-truth oriented boxes. (c) The box noise. (d) The angle noise. (e) The geometric noise. (f) The probability noise example.}
       \label{visual_contrastive_denoising}
    \end{figure}
}

\newcommand{\heatmap}{
    \begin{figure}[t]
      \centering
       \includegraphics[width=0.95\linewidth]{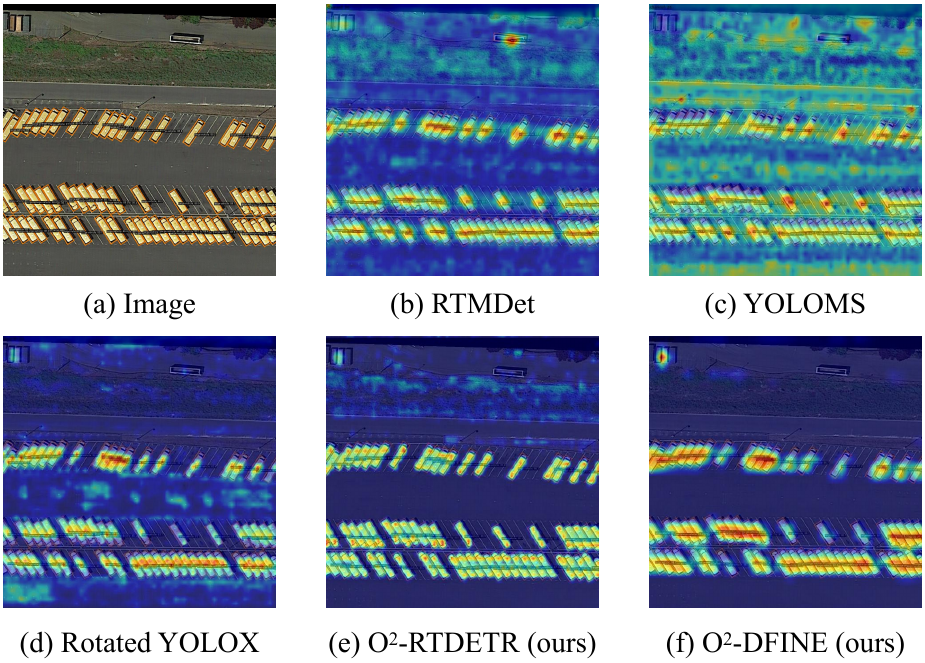}
       \caption{Visualization of backbone output feature maps. The heatmaps show that our method yields more concentrated and object-aligned activations compared to other methods.}
       \label{visual_heatmap}
    \end{figure}
}

\newcommand{\visualChamfer}{
\begin{figure}[t]
  \centering
       \includegraphics[width=0.95\linewidth]{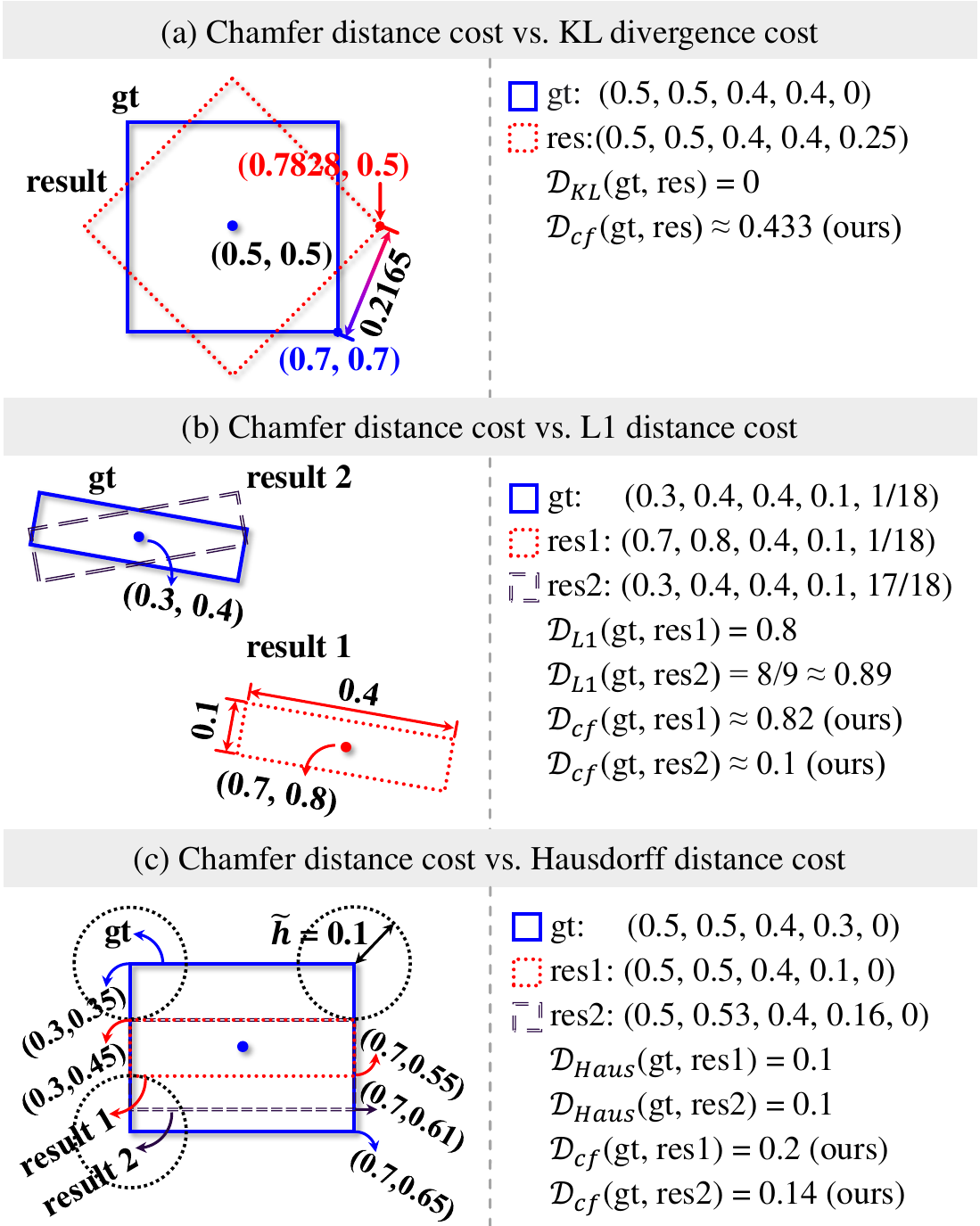}
  \caption{Comparison of Chamfer distance cost with KL divergence, L1, and Hausdorff distance costs. The centers, sizes, and angles are normalized.}
  \label{visual_chamfer}
  \vspace{-0.5em}
\end{figure}

}

\newcommand{\visualRoatedAttn}{
    \begin{figure}[t]
      \centering
       \includegraphics[width=0.95\linewidth]{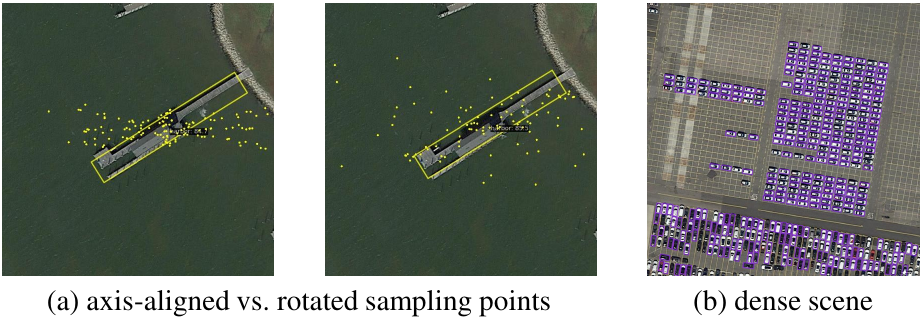}
       \caption{(a) Sampling points in cross attention. $\textcolor[rgb]{1,1,0}{\blacksquare}$ denotes harbors. (b) Suboptimal detection results in dense scenes. $\textcolor[rgb]{0.6, 0.2, 0.8}{\blacksquare}$ denotes cars.}
       \label{visual_rotated_attn}
    \end{figure}
}

\newcommand{\visualAngleDistRefine}{
    \begin{figure*}[t]
      \centering
       \includegraphics[width=\textwidth]{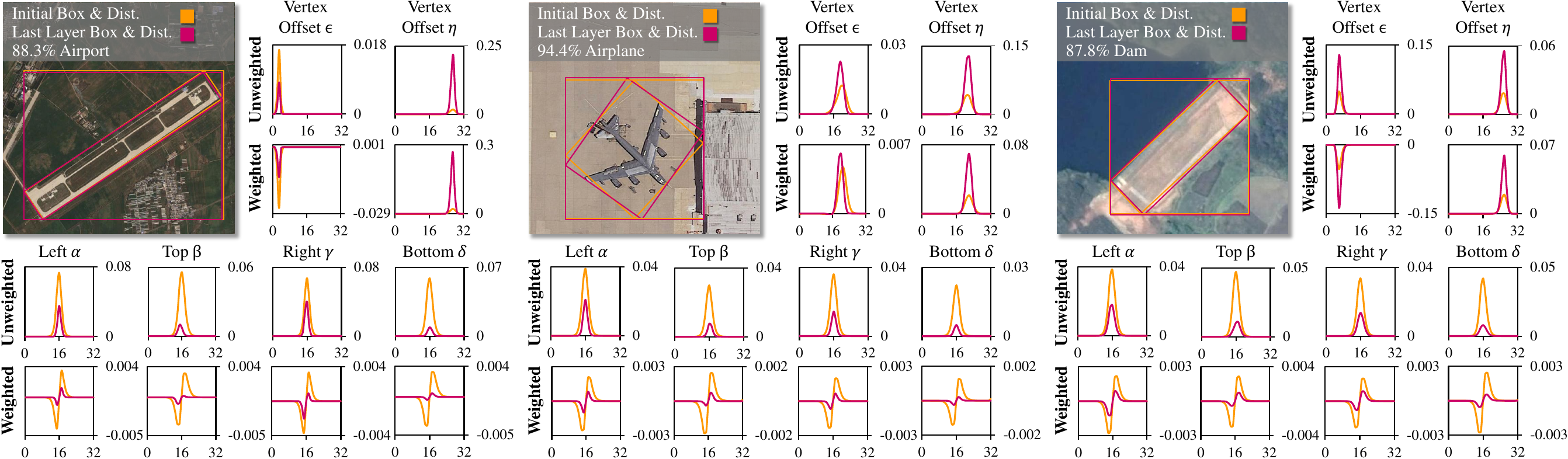}
       \caption{Analysis of Angle Distribution Refinement, including initial and refined oriented boxes, external rectangles, vertex offsets, with unweighted and weighted distributions.}
       \label{visual_angle_dist_refine}
    \end{figure*}
}

\newcommand{\visualresults}{
    \begin{figure*}[t]
      \centering
       \includegraphics[width=0.95\textwidth]{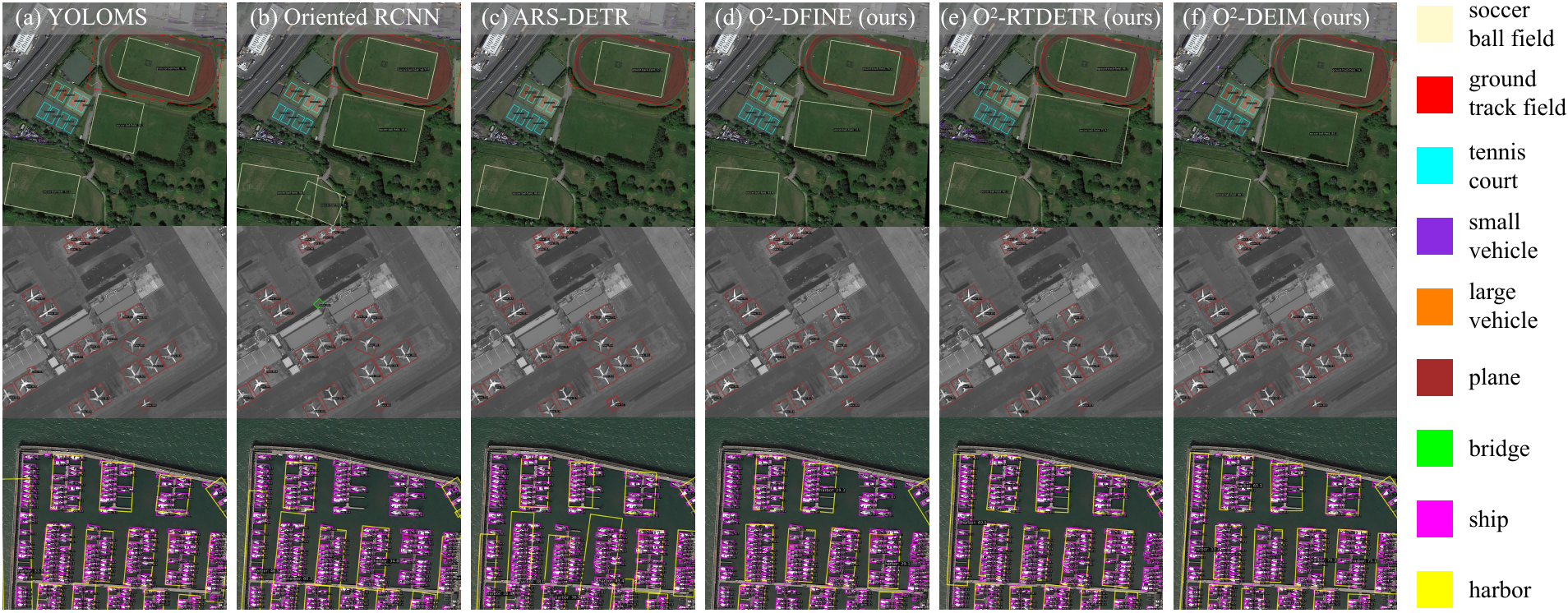}
       \caption{Qualitative comparison between our method and other detectors on challenging scenarios, including large-scale objects (1st row), low-light conditions (2nd row), and densely distributed objects (3rd row). The confidence threshold is set to 0.3}
       \label{visual_results}
    \end{figure*}
}

\newcommand{\AppendixvisualRoatedAttn}{
    \begin{figure}[t]
      \centering
       \includegraphics[width=0.95\linewidth]{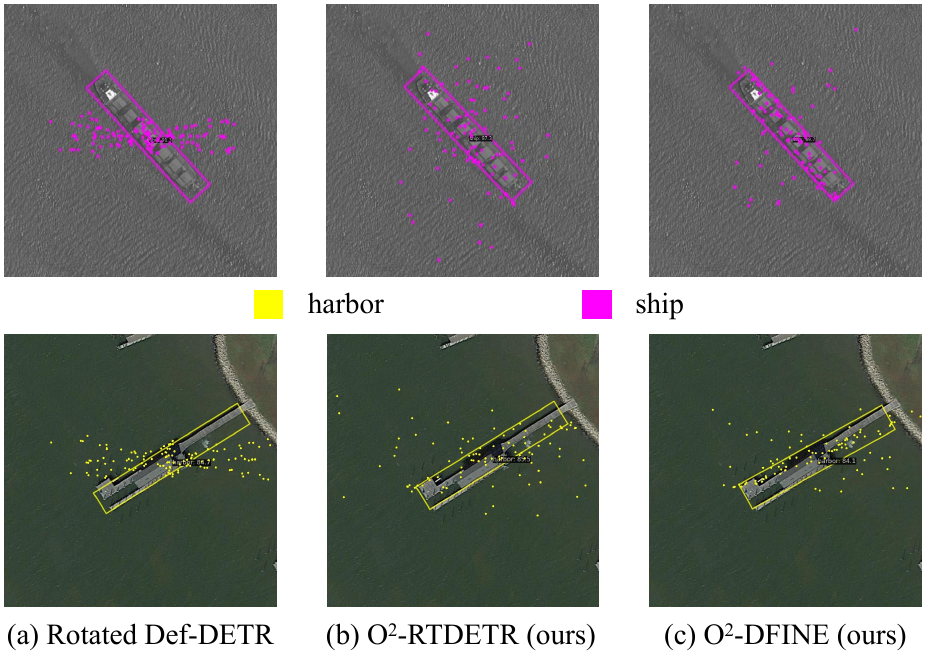}
       \caption{Visualization of sampling points in cross attention (a) Sampling points in Rotated Deformable DETR. (b) Rotated sampling points in O$^2$-RTDETR. (c) Rotated uneven sampling points in O$^2$-DFINE.}
       \label{appendix_visual_rotated_attn}
    \end{figure}
}

\begin{document}

\title{Real-Time Oriented Object Detection Transformer in Remote Sensing Images}

\author{
Zeyu Ding\textsuperscript{~\orcidlink{0009-0005-9614-7131}}, 
Yong Zhou\textsuperscript{~\orcidlink{0000-0001-6207-0299}}, 
Jiaqi Zhao\textsuperscript{~\orcidlink{0000-0002-3564-5090}},~\IEEEmembership{Member,~IEEE}, 
Wen-Liang Du\textsuperscript{~\orcidlink{0000-0002-9234-0912}}~\IEEEmembership{Member,~IEEE}, 
Xixi Li~\IEEEmembership{Member,~IEEE}, \\
Rui Yao\textsuperscript{~\orcidlink{0000-0003-2734-915X}},~\IEEEmembership{Member,~IEEE}, 
and Abdulmotaleb El Saddik\textsuperscript{~\orcidlink{0000-0002-7690-8547}},~\IEEEmembership{Fellow,~IEEE}

\thanks{Received 27 December 2025; revised 23 February 2026; accepted 3 March 2026.}

\thanks{Zeyu Ding, Yong Zhou, Jiaqi Zhao, Wen-Liang Du, Xixi Li, and Rui Yao are with the School of Computer Science and Technology / School of Artificial Intelligence, China University of Mining and Technology, Xuzhou 221116, China, and with the Mine Digitization Engineering Research Center of the Ministry of Education, Xuzhou 221116, China, and also with the Jiangsu Provincial Industrial Technology Engineering Center for Intelligent Sensing and Emergency IoT in Underground Space, Xuzhou 221116, China (e-mails: \{dingzeyu, yzhou, jiaqizhao, wldu, xixil, ruiyao\}@cumt.edu.cn).}

\thanks{Abdulmotaleb El Saddik is with the School of Electrical Engineering and Computer Science, University of Ottawa, Ottawa, ON K1N 6N5, Canada (e-mail: elsaddik@uottawa.ca).}

\thanks{Digital Object Identifier 10.1109/TGRS.2026.3671683}
}

% The paper headers 
\markboth{IEEE TRANSACTIONS ON GEOSCIENCE AND REMOTE SENSING}%
{DING et al.: REAL-TIME ORIENTED OBJECT DETECTION TRANSFORMER IN REMOTE SENSING IMAGES}

\maketitle

\begin{abstract}
Recent real-time detection transformers have gained popularity due to their simplicity and efficiency. However, these detectors do not explicitly model object rotation, especially in remote sensing imagery where objects appear at arbitrary angles, leading to challenges in angle representation, matching cost, and training stability.
In this paper, we propose a real-time oriented object detection transformer, the first real-time end-to-end oriented object detector to the best of our knowledge, that addresses the above issues. 
Specifically, angle distribution refinement is proposed to reformulate angle regression as an iterative refinement of probability distributions, thereby capturing the uncertainty of object rotation and providing a more fine-grained angle representation.
Then, we incorporate a Chamfer distance cost into bipartite matching, measuring box distance via vertex sets, enabling more accurate geometric alignment and eliminating ambiguous matches.
Moreover, we propose oriented contrastive denoising to stabilize training and analyze four noise modes. We observe that a ground truth can be assigned to different index queries across different decoder layers, and analyze this issue using the proposed instability metric.
We design a series of model variants and experiments to validate the proposed method. Notably, our O$^2$-DFINE-L, O$^2$-RTDETR-R50 and O$^2$-DEIM-R50 achieve 77.73\%/78.45\%/80.15\% AP$_{50}$ on DOTA1.0 and 132/119/119 FPS on the \textcolor[HTML]{76B900}{\raisebox{-0.15\height}{\simpleicon{nvidia}}} 2080ti GPU. Code is available at {\textcolor[HTML]{181717}{\raisebox{-0.25\height}{\simpleicon{github}}}} \url{https://github.com/wokaikaixinxin/ai4rs}.
\end{abstract}

\begin{IEEEkeywords}
Oriented object detection, detection transformer, real-time detector, remote sensing. 
\end{IEEEkeywords}

\section{Introduction}

\latency

Oriented object detection~\cite{orientedrcnn} in remote sensing images aims to classify objects into categories and localize them using rotated boxes aligned with their arbitrary orientations. Remote sensing applications typically require real-time detection with low latency. Existing real-time oriented object detection methods are primarily based on CNN models, with many adopting the YOLO framework~\cite{rtmdet,ppyoloer,yoloms} due to its reasonable trade-off between speed and accuracy. These approaches typically extend standard YOLO detectors by adding angle prediction branches to support rotated outputs. However, these models rely on post-processing steps such as rotated Non-Maximum Suppression (NMS)~\cite{nms}, which increases latency and introduces additional parameters, as shown in \figurename~\ref{latency}. Furthermore, in oriented object detection, the rotated NMS process requires a rotated IoU threshold, and the results are influenced by this threshold since oriented boxes are sensitive to IoU variations.

In recent years, end-to-end detection transformer (DETR) methods~\cite{detr} have gained wide attention in object detection due to their simple framework and the ability to remove hand-crafted components. 
DETR methods have been developed for oriented object detection, tackling challenges such as oriented object representation~\cite{orienteddetr}, attention mechanism design~\cite{arsdetr}, query formulation~\cite{rqformer}, and label assignment~\cite{rhino}.
However, the high computational cost often hinders their real-time performance, making them less suitable for real-time applications.

RT-DETR~\cite{rtdetr} emerged as the first real-time end-to-end detection transformer, achieving both efficiency and accuracy through a hybrid design and query selection mechanism. Following this pioneering work, several improved real-time Transformer-based detectors have been developed~\cite{rtdetrv2,rtdetrv3}. More recently, D-FINE~\cite{dfine} refines the regression process by modeling boxes as fine-grained distributions, enabling more precise localization. Methods like DEIM~\cite{deim} accelerate convergence through improved matching strategies. 
However, existing real-time detection transformers are primarily designed for horizontal object detection in natural scenes. They lack explicit support for oriented objects, which are common in remote sensing imagery.

Objects often appear at arbitrary orientations in remote sensing images, introducing geometric rotation transformations that pose several key challenges for real-time end-to-end detection transformers.
\ding{182} \textbf{\textit{Angle representation.}} Most detectors predict angles by regressing a radian value, treating rotation as a precise value modeled by a Dirac delta distribution~\cite{rtmdet}. This approach fails to capture the uncertainty in object rotation. Consequently, models rely on L1 or rotated IoU losses, which offer limited supervision for fine-grained angle adjustment. This makes optimization sensitive to minor angular errors and hinders convergence. While techniques such as DFL~\cite{ppyoloer} utilize distributions to represent uncertainty, they are still constrained by coarse localization and anchor dependency.
\ding{183} \textbf{\textit{Matching cost.}} Conventional distance metrics often introduce geometric ambiguity during bipartite matching. For instance, L1 distance may assign a lower cost to a spatially distant prediction that happens to share the same orientation, while penalizing a well-aligned but slightly rotated box. Similarly, KL divergence~\cite{kld} fails to distinguish between square boxes with identical shape and center, and Hausdorff distance~\cite{rhino} is overly sensitive to outliers.
\ding{184} \textbf{\textit{Training stability.}} We observe that a single ground-truth object can be matched to different index predicted queries across decoder layers during early training stages. The instability of bipartite graph matching causes inconsistent optimization goals. Hungarian matching exhibits unstable assignment due to the existence of blocking pairs~\cite{blockpair}. A slight change in the cost may cause a significant change in the matching result.

To address these issues, we propose a real-time end-to-end oriented object detection transformer for remote sensing images that introduces three key innovations for the challenges of angle representation, matching cost, and training stability. 
\textit{\textbf{First}}, we design \textbf{Angle Distribution Refinement}, which replaces direct angle regression with an iterative refinement of probability distributions, enabling the model to capture angular uncertainty and perform progressively finer residual adjustments in a residual manner.
\textit{\textbf{Second}}, we incorporate \textbf{Chamfer Distance Cost} into bipartite matching by measuring the distance between oriented boxes via their vertices, which alleviates ambiguous matches caused by traditional distance costs and ensures that only spatially and rotationally aligned predictions are favored.
\textit{\textbf{Third}}, we introduce \textbf{Oriented Contrastive Denoising}, a training strategy that injects noise into oriented boxes to generate positive and negative query pairs across four modes. We propose an instability metric to verify that this mechanism stabilizes label assignment.
We develop a family of detectors and perform extensive experiments on four datasets. Our O$^2$-DFINE-L, O$^2$-RTDETR-R50, and O$^2$-DEIM-R50 achieve 77.73\%/78.45\%/80.15\% AP$_{50}$ on DOTA1.0 and 132/119/119 FPS on the 2080ti GPU.

The contributions of this work are summarized as follows:
\begin{enumerate}
\renewcommand{\labelenumi}{\arabic{enumi})}
\item To model the uncertainty in object rotation angles, we introduce Angle Distribution Refinement, which provides a fine-grained representation and enables progressively finer adjustments.
\item To alleviate ambiguous matches caused by suboptimal distance costs, we incorporate the Chamfer Distance Cost into bipartite matching, enabling more accurate geometric alignment.
\item To mitigate training instability in label assignment, we propose Oriented Contrastive Denoising with four noise injection modes. We further introduce an instability metric to verify its effectiveness.
\item As far as we know, our method is the first real-time oriented object detection transformer, with variants including O$^2$-RTDETR, O$^2$-DFINE, and O$^2$-DEIM. Experiments demonstrate that the method achieves competitive performance with real-time inference speed.
\end{enumerate}

\section{Related Work}

\subsection{Oriented Object Detection in Remote Sensing}
Oriented object detection in remote sensing imagery aims to localize objects with arbitrary orientations. Existing methods can be broadly categorized into CNN-based, transformer-based, and random-initialization approaches according to how object candidates are generated. 
\underline{CNN-based methods} include both two-stage~\cite{orientedrcnn,redet} and single-stage~\cite{messdet,dcfl} detectors. Two-stage methods generate region proposals that are subsequently refined, while single-stage methods directly predict object classes and oriented bounding boxes from densely placed anchors in a single forward pass. 
\underline{Transformer-based methods} formulate oriented object detection as a set prediction problem using a fixed number of learnable queries~\cite{orienteddetr,arsdetr}. These methods directly predict oriented bounding boxes in an end-to-end manner, eliminating the need for anchors or proposal generation.
\underline{Random-initialization methods} represent object candidates using randomly initialized primitives. These methods refine Gaussians into oriented boxes through Gaussian splatting~\cite{gsdet} or denoising diffusion~\cite{rediffdet}, and typically support a dynamic number of candidates and enable flexible inference.

In contrast to existing CNN-based~\cite{msc,friou,wsod,10473137,guo2025multi} and random-initialization methods, our approach is fully end-to-end and avoids handcrafted post-processing such as non-maximum suppression. Compared with prior transformer-based detectors, our method further achieves real-time inference.

\subsection{Detection Transformer in Oriented Object Detection}

Existing detection transformers for oriented object detection can be broadly categorized by their core technical designs, including oriented object representation, rotation-aware attention, query formulation, and label assignment.
Some works focus on \underline{oriented object representations} to mitigate angle ambiguity, such as point-based~\cite{orienteddetr} or oriented-box-based~\cite{ro2detr} query formulations, or decoupled angle modeling to reduce error propagation~\cite{orienteddino}. 
Another line of research enhances \underline{attention mechanisms} with rotation awareness, for example, by rotating sampling locations~\cite{arsdetr}, introducing rotation-equivariant feature extraction~\cite{ro2detr}, or designing geometry-aware self- and cross-attention modules~\cite{orientedformer} to better capture oriented object structures.
\underline{Query design} has also been explored to improve efficiency and suppress redundancy, including representing objects as point sets, generating selective distinct queries~\cite{rqformer}, and encouraging query diversity during training~\cite{orienteddino}.
\underline{Label assignment strategies} are further adapted to aerial imagery, addressing unbalanced object distributions and redundant predictions through dynamic matching schemes~\cite{ro2detr}, geometry-aware matching costs~\cite{rhino}, or high-quality sample selection~\cite{emo2detr}.

In contrast to these methods, our method represents the first real-time detection transformer for oriented object detection.

\differentConfThre
\NMSAP

\subsection{Real-time Object Detectors}
Real-time object detection has remained a key focus area within computer vision. Among the most influential frameworks in this area is the \underline{YOLO series}~\cite{yolov1}. Early YOLO models established the one-stage detection paradigm for fast inference, while later variants improved the speed–accuracy trade-off through enhanced multi-scale feature representations~\cite{yolov6}, decoupled detection heads~\cite{yolov8}, efficient backbones~\cite{yolov9}, and attention modules~\cite{yolov12}.

Despite their efficiency, most YOLO-based detectors rely on NMS to remove duplicate predictions, which introduces additional latency and may lead to unstable results in dense scenes, such as remote sensing imagery. This limitation has motivated the development of end-to-end real-time detectors. RT-DETR~\cite{rtdetr} emerged as the first end-to-end \underline{real-time detection transformer} by introducing efficient encoder design and query selection. Subsequent works further improved localization accuracy and training efficiency by refining regression representations and enhancing matching-based supervision~\cite{dfine,deim}.

Unlike existing real-time detection transformers, our method explicitly supports oriented object detection and robustly handles arbitrarily rotated objects commonly found in remote sensing imagery.

\section{Analysis of rotated NMS for Detectors}

\NMSExeTime

Rotated Non-Maximum Suppression (NMS) is a post-processing algorithm designed to remove overlapping oriented boxes in non-end-to-end oriented object detectors. Rotated NMS requires two thresholds: a rotated IoU threshold and a confidence threshold. It first filters out predicted oriented boxes whose scores fall below the confidence threshold. Then, whenever the rotated IoU between two oriented boxes exceeds the rotated IoU threshold, the box with the lower score is discarded. This process continues until all boxes have been processed. The latency of rotated NMS is influenced by the two thresholds and the number of oriented boxes.

To analyze these effects, we count how many oriented boxes remain after filtering the predictions with different confidence thresholds (0.005 to 0.25) on the DOTA-v1.0 \texttt{trainval} set. We adopt anchor-free detectors, RTMDet-s~\cite{rtmdet} and YOLOX-s~\cite{yolox}, and the anchor-based YOLOv5-s~\cite{yolov5}. The results in \figurename~\ref{differentConfThre} show that rotated NMS is highly sensitive to the confidence threshold. As the confidence threshold increases, more oriented boxes are removed, leaving fewer candidates for the rotated IoU computation. In addition, we evaluate the AP$_{50}$ of RTMDet-s on the DOTA-v1.0 \texttt{test} set under different thresholds, as shown in Table~\ref{NMSAP}. The results show that the detection accuracy is influenced by both thresholds.

Furthermore, we test the execution time of the rotated NMS operation under different numbers of non-overlapping oriented boxes, with a single class and all scores set to 100\%, as shown in \figurename~\ref{NMSExeTime}. The test is conducted on an \textcolor[HTML]{76B900}{\raisebox{-0.15\height}{\simpleicon{nvidia}}} 2080ti GPU using TensorRT FP16 with NVIDIA \texttt{trtexec} and \href{https://developer.nvidia.com/nsight-systems}{\textcolor{black}{Nsight Systems}}. The rotated NMS operation adopted here is the \texttt{RotatedNMSPlugin} \href{https://github.com/laugh12321/TensorRT-YOLO}{\textcolor[HTML]{181717}{\raisebox{-0.25\height}{\simpleicon{github}}}}, which is provided as a TensorRT static plugin. Its key kernels include \texttt{RotatedNMS}, \texttt{RotatedNMSFilter}, and \texttt{RadixSort}.  As the number of oriented boxes increases, the execution time of \texttt{RotatedNMS} grows significantly, whereas the overhead of the other two kernels remains negligible.

\section{Method}

\angleDistribution

\subsection{Overview}

% Our model is composed of a backbone, a transformer encoder, a decoder, and a prediction head. 

Our approach is centered on addressing the challenges caused by rotation transformations, including angle representation (Sec.~\ref{sec:ADR}), matching cost (Sec.~\ref{sec:chamfer}), and training stability (Sec.~\ref{sec:denoising}). The training procedure is shown in Algorithm \ref{algo}. The model architecture is introduced below.

\textit{Oriented boxes definition.} We represent an oriented bounding box as $(c_x, c_y, w, h, \theta)$, where $(c_x, c_y)$ denotes the box center, $w$ and $h$ are the width and height, and $\theta \in [0, \pi)$ represents the rotation angle relative to the x-axis.

\textit{Backbone with channel mapper.} The O$^2$-RTDETR and O$^2$-DEIM adopts the ResNet~\cite{resnetd} series as the backbone, while O$^2$-DFINE employs the HGNetv2~\cite{paddlepaddle} series. Both backbones output three feature maps with downsampling ratios of 8$\times$, 16$\times$, and 32$\times$ relative to the input image. To ensure consistent feature dimensions for the encoder, each feature map is projected to 256 channels using a $1\times1$ convolution layer.

\textit{Transformer encoder with query selection.} The encoder follows the hybrid encoder~\cite{rtdetr}, decoupling intra-scale interaction and cross-scale fusion. An attention-based module performs self-attention on the highest-level feature map, while a CNN-based module fuses features across scales efficiently. The uncertainty-minimal query selection~\cite{rtdetr} is employed to select the top-$K$ (default 300) queries, and the corresponding oriented boxes are used as references for the decoder.

\textit{Transformer decoder.} The decoder consists of multiple stacked decoder layers, each containing self-attention, cross-attention, and feed-forward networks. In cross-attention, the sampling points are rotated around the box center according to the predicted box angle. 
%In D-FINE~\cite{dfine}, the cross-attention performs uneven sampling across multiple feature scales. We extend this idea by rotating these unevenly sampled points according to the box orientation, forming the cross-attention used in O$^2$-DFINE.

\begin{algorithm}[h]
    \caption{Training procedure pseudocode.}
   \label{algo}
\definecolor{codeblue}{rgb}{0.29, 0.53, 0.91}
\lstset{
   basicstyle=\footnotesize\ttfamily,        % 适中的字体
  commentstyle=\color{codeblue}\itshape,     % 注释为斜体并着色
  keywordstyle=\bfseries,                    % 关键词加粗
  stringstyle=\color{orange},                % 字符串用橙色
  breaklines=true,                           % 自动换行
  backgroundcolor=\color{white},             % 背景色
  showstringspaces=false,                    % 不显示字符串中的空格
  aboveskip=0pt,                             % 减少上边距
  belowskip=0pt,                             % 减少下边距
}
\begin{lstlisting}[language=python, mathescape, breaklines=true]
# adr - head with angle distribution refinement
# ocd - oriented contrastive denoising
# cdc - matching with Chamfer distance cost
prediction = adr(transformer(backbone(image)))
noise = ocd(gt) # gt - ground truth
pairs = cdc({prediction, gt}, {noise, gt})
loss(pairs)
\end{lstlisting}
\end{algorithm}

\subsection{Angle Distribution Refinement}\label{sec:ADR}

Angle Distribution Refinement (ADR) iteratively optimizes the fine-grained distributions of the rotation angle $\theta$ and the parameters $(c_x,c_y,w,h)$ across decoder layers, as shown in \figurename~\ref{angleDistribution}. Instead of directly modeling the angle $\theta$ in the angular space, ADR decomposes the prediction process into two stages: (1) refining the external rectangle of an oriented box, and (2) adjusting the vertex offsets along its edges to derive the final oriented box. This indirect formulation ensures that both the angle $\theta$ and the parameters $(c_x, c_y, w, h)$ are optimized within a unified metric space, thus avoiding inconsistency.
The first decoder layer predicts initial oriented boxes through a traditional oriented box regression head and initial distributions through an O$^2$-DFINE head. Each oriented box is associated with six distributions: four for the edges of its external rectangle and two for the offsets along these edges. The subsequent layers iteratively refine both types of distributions in a residual manner. The refined distributions are used to adjust the external rectangle and its edge offsets, jointly refining the angle $\theta$ and the parameters $(c_x, c_y, w, h)$. 

\textit{Vertex offset.}
The initial vertex offsets $\mathcal{D}_o^0 = (\epsilon, \eta)$ are defined such that $\epsilon$ measures the distance from the top vertex of the oriented box to the top-right corner of its external rectangle, while $\eta$ measures the distance from the rightmost vertex to the bottom-right corner.
The vertex offsets $(\epsilon, \eta)$ characterize how the oriented box is positioned within its external rectangle.
For the $l$-th decoder layer, the vertex offsets $\mathcal{D}_o^l=(\epsilon^l,\eta^l)$ are updated in a residual manner as:
\begin{equation}
\mathcal{D}_o^l = \mathcal{D}_o^0 + (W_r, H_r) \cdot \sum_{n=0}^{N} \mathcal{A}(n)\mathcal{P}_o^{l}(n), 
\end{equation}
where the decoder layer $l\in\{1,2,...,L\}$. $\mathcal{P}_o^l(n)=\{\mathcal{P}_\epsilon^l(n),\mathcal{P}_\eta^l(n)\}$ denotes two distributions corresponding to vertex offset. Each distribution predicts the likelihood of candidate offset values. These candidate offsets are discretized into $N+1$ bins indexed by $n$, with each bin representing one possible offset value. The weighting function $\mathcal{A}(n)$ maps each bin index $n$ to its corresponding offset.

\textit{External rectangle.} Let $r^0 = (c_x, c_y, W_r, H_r)$ denote the initial external rectangle converted from the initial oriented boxes, where $(c_x, c_y)$ is the center and $(W_r, H_r)$ represent the width and height of the external rectangle. Let $\mathcal{D}_r^0 = (\alpha,\beta,\gamma,\delta)$ denote the distances from the center to the left, top, right, and bottom edges of the external rectangle, respectively. 
For the $l$-th decoder layer, the refined edge distances $\mathcal{D}_r^l = (\alpha^l, \beta^l, \gamma^l, \delta^l)$ are updated in a residual manner as:
\begin{equation}
\mathcal{D}_r^l = \mathcal{D}_r^0 + (W_r, H_r, W_r, H_r) \cdot \sum_{n=0}^{N} \mathcal{A}(n)\mathcal{P}_r^{l}(n), 
\end{equation}
where $\mathcal{P}_r^l(n)=\{\mathcal{P}_\alpha^l(n),\mathcal{P}_\beta^l(n),\mathcal{P}_\gamma^l(n),\mathcal{P}_\delta^l(n)\}$ denotes four distributions corresponding to the edges of the external rectangle.

\charmferDistance

\textit{Refined probability distribution.}
ADR refines the probability distributions in a residual manner, formulated as:
\begin{equation}
\mathcal{P}^l_\cdot(n)=\text{Softmax}(\text{logits}^{l-1}(n) + \Delta \text{logits}^l(n)).
\end{equation}
The $\text{logits}^{l-1}(n)$ represent the confidence of the previous decoder layer in each discretized bin of the distance values for the vertex offsets and external rectangle.
The current decoder layer outputs residual logits $\Delta \text{logits}^l(n)$, and the updated logits are computed as $\text{logits}^l(n)=\text{logits}^{l-1}(n)+\Delta \text{logits}^l(n)$.
These updated logits are then transformed into refined probability distributions via softmax normalization.

\textit{Weighting function.}
The weighting function is defined as:
\begin{equation}\label{weight_fun}
\mathcal{A}(n) = \operatorname{sgn}(n-\frac{N}{2})\cdot 
\begin{cases} 
2a, \, n\in\{0,N\},\\
c\Big((1+\frac{a}{c})^{\frac{2\left| n-\frac{N}{2} \right|}{N-2}}-1\Big), \, \text{other}.
\end{cases} 
\end{equation}
where $n\in\{0,1,\dots,N\}$ and $\operatorname{sgn}(\cdot)$ denotes the sign function. 
$a$ and $c$ are hyper-parameters that control the curvature and bounds of the weighting function. 
When the predicted box is close to the target, the smooth variation of $\mathcal{A}(n)$ enables fine-grained adjustments. 
In contrast, for inaccurate predictions, the sharp changes near the boundaries of $\mathcal{A}(n)$ allow larger corrective updates.

\textit{Loss.}
The Fine-Grained Localization loss~\cite{dfine} is adopted and applied to both vertex offsets and external rectangle.

\subsection{Chamfer distance cost for bipartite matching}\label{sec:chamfer}

In detection transformers, label assignment is formulated as a bipartite matching problem solved by the Hungarian algorithm. The matching cost typically consists of three terms: distance, IoU, and classification costs. The distance cost measures the spatial alignment between predictions and ground truths. This section focuses on the distance cost, analyzing the limitations of commonly used formulations and introducing the Chamfer distance cost as a more robust and geometry-consistent alternative, as shown in \figurename~\ref{charmferDistance}.

\textit{Chamfer distance.}
Chamfer distance is a metric used to measure the similarity between two point sets by calculating the sum of the squared distances from each point in one cloud to its nearest neighbor in the other cloud. Given two point sets $\mathcal{S}$ and $\mathcal{T}$, it is defined as:
\begin{equation}\label{eq:chamferdistance}
\mathcal{D}_{\text{cf}}(\mathcal{S}, \mathcal{T}) =
\frac{1}{|\mathcal{S}|}\sum_{s \in \mathcal{S}} \min_{t \in \mathcal{T}} \|s - t\|_2^2 +
\frac{1}{|\mathcal{T}|}\sum_{t \in \mathcal{T}} \min_{s \in \mathcal{S}} \|t - s\|_2^2.
\end{equation}
The first term measures the distance from each point in set $\mathcal{S}$ to its nearest neighbor in $\mathcal{T}$, while the second term performs the reverse, ensuring symmetry by measuring distances from $\mathcal{T}$ to $\mathcal{S}$.

\contrastiveDenoise

\textit{Chamfer distance cost.}
Both the predicted and ground-truth oriented boxes are converted into point sets composed of their four corner vertices, denoted as $\mathcal{S}_{pred}$ and $\mathcal{S}_{gt}$. 
Given an oriented box parameterized by $(c_x, c_y, w, h, \theta)$, its four vertices $\{v_1,v_2,v_3,v_4\}$ can be derived as:
\begin{equation}\label{eq:rbox2vertex}
v_i = (c_x, c_y) \pm \frac{w}{2}(\cos\theta, \sin\theta) \pm \frac{h}{2}(-\sin\theta, \cos\theta),
\end{equation}
where $i\in\{1,2,3,4\}$. The ordering of vertices is not constrained, as the Chamfer distance is permutation-invariant with respect to point correspondence. The Chamfer distance cost between the two sets is computed as:
\begin{equation}
\mathcal{C}_{\text{cf}} = \mathcal{D}_{\text{cf}}(\mathcal{S}_{pred}, \mathcal{S}_{gt}).
\end{equation}
During bipartite matching, this Chamfer distance cost serves as the distance term in the overall matching cost. The point-based Chamfer distance cost preserves geometric alignment and mitigates matching ambiguity. The following discusses the limitations of conventional cost formulations and the advantages of our proposed method.

\textit{Geometric Ambiguity in L1 distance cost.}
The L1 distance measures absolute differences in center coordinates, widths, heights, and angles. Given a ground-truth box $(c_x, c_y, w, h, \theta)$ and a predicted one $(c_x+\Delta c_x,c_y+\Delta c_y, w+\Delta w, h+\Delta h, \theta+\Delta \theta)$, the L1 distance between them is defined as:
\begin{equation}
\mathcal{D}_\text{L1}=\left |\Delta c_x \right|+ \left |\Delta c_y \right|+ \left | \Delta w \right|+ \left | \Delta h \right|+ \left | \Delta \theta \right|.
\end{equation}
\figurename~\ref{charmferDistance}(c) illustrates the geometric ambiguity of the L1 distance. A predicted box $(c_x+\Delta c_x, c_y+\Delta c_y, w, h, \theta)$ that is far away from the ground truth and has zero IoU can still obtain a smaller L1 cost than another prediction $(c_x, c_y, w, h, \theta+\Delta \theta)$ that is perfectly aligned in position and size but differs only by a rotation. The geometric relationships are expressed as:
\begin{gather}
\sqrt{\Delta c_x^2+\Delta c_y^2} \cos(\theta+\arctan \frac{\Delta c_x}{\Delta c_y})>h, \label{l1distance1}\\
\left | \Delta\theta \right | >\left |  \Delta c_x\right | +\left | \Delta c_y \right |. \label{l1distance2}
\end{gather}
When both Eq.~\ref{l1distance1} and Eq.~\ref{l1distance2} are satisfied, the distant prediction $(c_x+\Delta c_x, c_y+\Delta c_y, w, h, \theta)$ obtains a lower L1 cost than the prediction that is rotated by $\Delta \theta$. In contrast, our Chamfer distance assigns a higher cost to the distant prediction and avoids this geometric ambiguity.

\textit{Square problem in KL divergence cost.}
The Kullback-Leibler divergence treats oriented boxes as probability distributions and measures the distance between them~\cite{kld}, as shown in \figurename~\ref{charmferDistance}(b). For a square box $(c_x, c_y, w, w, \theta)$, the means of its distribution is $\boldsymbol{\mu}=(c_x,c_y)^\top$ and the variance matrix is $\boldsymbol{\Sigma} = \bigl( \begin{smallmatrix} w^2 / 4 & 0 \\ 0 & w^2 / 4 \end{smallmatrix} \bigr)$. Two square boxes that differ only in angle have the same mean $\boldsymbol{\mu}$ and variance $\boldsymbol{\Sigma}$, and the KL divergence between them is computed as follows:
\begin{equation}
\mathcal{D}_{\text{KL}} = \frac{1}{2}(\boldsymbol{\mu}-\boldsymbol{\mu})^\top\boldsymbol{\Sigma}^{-1}(\boldsymbol{\mu}-\boldsymbol{\mu})+
\frac{1}{2}\mathrm{tr}(\boldsymbol{\Sigma}^{-1}\boldsymbol{\Sigma})+
\frac{1}{2}\ln\frac{\left|\boldsymbol{\Sigma}\right|}{\left|\boldsymbol{\Sigma}\right|}-1.
\end{equation}
When two rotated square boxes differ only in their angles, their KL divergence becomes 0, making it unable to distinguish between them. In contrast, the Chamfer distance can still provide a meaningful measure of their difference.

\textit{Outlier sensitivity in Hausdorff distance cost.}
The Hausdorff distance measures how far two point sets are from each other by focusing on their worst-case deviation~\cite{rhino}. Given two point sets $\mathcal{S}$ and $\mathcal{T}$, the Hausdorff distance is defined as the maximum distance from a point in one set to its nearest neighbor in the other set:
\begin{equation}
\mathcal{D}_{\text{Haus}} = \max \left \{ \max_{s\in\mathcal{S}} \min_{t\in\mathcal{T}}\left \| s-t \right \|_2, \max_{t\in\mathcal{T}} \min_{s\in\mathcal{S}}\left \| t-s \right \|_2 \right \}.
\end{equation}
The first term measures the greatest among all the smallest distances from points in $\mathcal{S}$ to points in $\mathcal{T}$, while the second term measures the reverse. By taking the maximum of these two, the Hausdorff distance captures the largest mismatch between the two sets. \figurename~\ref{charmferDistance}(d) illustrates a case where the Hausdorff distances $\tilde{h}$ ($\tilde{h}<\frac{h}{2}$) between the predicted and ground-truth boxes are all the same. In contrast, the Chamfer distance is less sensitive to outliers.

\subsection{Oriented Contrastive Denoising}\label{sec:denoising}

The Oriented Contrastive Denoising (OCD) is specifically designed for oriented boxes as shown in \figurename~\ref{contrastiveDenoise}. OCD generates paired positive and negative queries with noise for each ground-truth oriented box, where the positive queries receive slight noise, and the negative ones receive larger noise. Through this denoising process, the decoder learns to recover accurate box geometry and stabilize training. We explore four noise modes across different dimensions of the oriented boxes. The instability metric is proposed to verify effectiveness.

\textit{Box noise.}
In this setting, the angle $\theta$ is kept unchanged. Instead of noise addition into $(c_x,c_y,w,h)$ directly, we first convert the oriented box into $(x_1,y_1,x_2,y_2)$, where $x_1,x_2 = c_x \pm w/2,$ and $ y_1,y_2 = c_y \pm h/2$. Noise is then added to each coordinate of the vertices, yielding $(x_1+\Delta x_1,y_1+\Delta y_1)$ and $(x_2+\Delta x_2,y_2+\Delta y_2)$. The noise magnitude is controlled by two hyperparameters $\lambda_1$ and $\lambda_2$ ($\lambda_1 < \lambda_2$). For the positive query, we apply small noise to each vertex. For the vertex $(x_1,y_1)$, the noise ranges are defined as:
\begin{equation}
  \Delta x_{1}^{\text{pos}} \in [-\frac{\lambda_1 w}{2},\frac{\lambda_1 w}{2}), \quad
  \Delta y_{1}^{\text{pos}} \in [-\frac{\lambda_1 h}{2},\frac{\lambda_1 h}{2}),
\end{equation}
and the same formulation is applied symmetrically to $(x_2,y_2)$. This produces a slightly noised oriented box whose shape and location remain close to the ground truth. In contrast, the negative query receives larger noise:
\begin{equation}
  \Delta x_{1}^{\text{neg}} \in 
  [-\frac{\lambda_2 w}{2}, -\frac{\lambda_1 w}{2}) \cup
  [\frac{\lambda_1 w}{2}, \frac{\lambda_2 w}{2}),
\end{equation}
and similarly for $\Delta y_{1}^{\text{neg}}, \Delta x_{2}^{\text{neg}}, \Delta y_{2}^{\text{neg}}$. These larger vertex shifts produce a significantly distorted box that differs noticeably from the ground truth.

\instability

\textit{Angle noise.} 
In this setting, noise is only injected into the angle $\theta$ while the parameters $(c_x,c_y,w,h)$ remain unchanged. For each ground-truth oriented box, we generate paired positive and negative queries by injecting angular noise, controlled by two hyperparameters $\lambda_3$ and $\lambda_4$ ($\lambda_3 < \lambda_4$). Specifically, the positive query receives a small random noise:
\begin{equation}
\Delta\theta^{\text{pos}}\in[-\frac{\lambda_3\theta}{18}, \frac{\lambda_3\theta}{18}),
\end{equation}
which produces an oriented box whose orientation remains close to the ground truth. In contrast, the negative query is noised with a larger angular deviation:
\begin{equation}
\Delta\theta^{\text{neg}}\in[-\frac{\lambda_4\theta}{18}, \frac{\lambda_3\theta}{18})\cup[\frac{\lambda_3\theta}{18}, \frac{\lambda_4\theta}{18}),
\end{equation}
resulting in a clearly misaligned orientation.

\textit{Geometric noise.}
The geometric noise setting combines the box noise and angle noise, jointly applying noise to both the box geometry and the rotation angle. Specifically, the parameters $(c_x,c_y,w,h)$ are first converted into $(x_1,y_1,x_2,y_2)$, and geometric noise is added to the vertices following the box noise strategy. At the same time, the angle $\theta$ is perturbed following the angle noise formulation.

\textit{Probability noise.}
Inspired by ReDiffDet~\cite{rediffdet}, each oriented box is represented as a Gaussian distribution $\mathcal{N}(\mathbf{0}, \boldsymbol{\Sigma})$, where the box center is placed at the origin, and the covariance is defined as $\boldsymbol{\Sigma} = \mathbf{R} \mathbf{\Lambda}
 \mathbf{R}^\top$, where $\mathbf{R}=\bigl( \begin{smallmatrix} \cos \theta & -\sin \theta \\ \sin & \cos \theta \end{smallmatrix} \bigr)$ and $\mathbf{\Lambda} = \operatorname{diag}((\frac{w}{max(w,h)})^2/4, (\frac{h}{max(w,h)})^2/4)$. A standard Gaussian noise $\mathcal{N}(\mathbf{0},\mathbf{I})$ is added to the distribution. Using the additive property of Gaussian variables, we obtain:
\begin{equation}
  \sqrt{1-\lambda}\mathcal{N}(\mathbf{0}, \boldsymbol{\Sigma}) + \sqrt{\lambda} \mathcal{N}(\mathbf{0}, \mathbf{I}) = \mathcal{N}(\mathbf{0}, (1-\lambda)\boldsymbol{\Sigma}+\lambda\mathbf{I}).
\end{equation}
Two coefficients $\lambda_5$ and $\lambda_6$ ($\lambda_5 < \lambda_6$) control the perturbation strength for positive and negative queries. The covariance of a positive query is sampled from $\boldsymbol{\Sigma}^{pos}\in[\boldsymbol{\Sigma},(1-\lambda_5)\boldsymbol{\Sigma}+\lambda_5\mathbf{I})$, while negative queries use $\boldsymbol{\Sigma}^{neg}\in[(1-\lambda_5)\boldsymbol{\Sigma}+\lambda_5\mathbf{I},(1-\lambda_6)\boldsymbol{\Sigma}+\lambda_6\mathbf{I})$. Finally, the noised oriented box is reconstructed from the perturbed covariance via eigenvalue decomposition.

\textit{Effectiveness.}
\figurename~\ref{contrastiveDenoise}(e) shows that a single ground-truth object can be matched to different index queries across decoder layers.
For a training image, we denote $M$ ground truth objects as $\{\text{gt}_0,\text{gt}_1, \dots ,\text{gt}_{M-1}\}$, and predicted objects from $l$-th decoder layer as $\{\text{query}^l_0,\text{query}^l_1,\dots,\text{query}^l_{K-1}\}$. After bipartite matching, we compute an index vector $\mathbf{G}^l=\{G^l_0,G^l_1,\dots,G^l_{M-1}\}$ to store the matching result of $l$-th layer as follows:
\begin{equation}
    G^l_m = k, \quad \text{if } \text{gt}_m \text{ matches } \text{query}^l_k.
\end{equation}
We define the instability for one training image as the difference among $\{\mathbf{G}^0,\mathbf{G}^1,\dots,\mathbf{G}^{L-1}\}$, which is calculated as:
\begin{equation}
  IS = (\sum_{m=0}^{M-1} G^0_m \oplus G^1_m \dots \oplus G^{L-1}_m)/M,
\end{equation}
where $\oplus$ is Exclusive OR. If the matching query indices change across layers, $IS$ will count them. \figurename~\ref{instability} shows a comparison of instability between rotated Deformable DETR and O$^2$-DFINE-L with different noise modes. We conduct this evaluation on the DOTA-v1.0 \texttt{trainval} set during training 1-12 epochs. The Oriented Contrastive Denoising effectively alleviates the instability of matching. 

\TableExpSetting

\section{Experiments}
\subsection{Datasets}
We conduct extensive experiments on four datasets DOTA-v1.0~\cite{dotav1.0}, DOTA-v1.5, DIOR-R~\cite{dior}, FAIR1M-v1.0~\cite{fair1m}.
DOTA-v1.0~\cite{dotav1.0} contains 1,869 images in the \texttt{trainval} set and 937 images in the \texttt{test} set, with 188,282 instances annotated across 15 common object categories. DOTA-v1.5 uses the same images and splits as DOTA-v1.0 but includes more small objects and a new class, resulting in 16 categories and 403,318 instances.
The DIOR-R~\cite{dior} dataset comprises 11,725 images in the \texttt{trainval} set and 11,738 in the \texttt{test} set, covering 20 object categories and 192,512 annotated instances.
The FAIR1M-v1.0~\cite{fair1m} dataset contains 15,266 high-resolution remote sensing images, encompasses 37 fine-grained object categories organized under 5 coarse categories, and provides over 1.02 million annotated instances.

\DOTAOneResults

\subsection{Implementation Details and Evaluation Metrics}

\textit{Implementation details.}
We conduct all the experiments on two \textcolor[HTML]{76B900}{\raisebox{-0.15\height}{\simpleicon{nvidia}}} 2080ti with a batch of 8 (4 images per GPU). Models are constructed based on AI4RS with Pytorch \textcolor[HTML]{EE4C2C}{\raisebox{-0.15\height}{\simpleicon{pytorch}}}. We optimize models with the AdamW optimizer. O$^2$-RTDETR and O$^2$-DEIM employ KLD loss $\mathcal{L}_{kld}$~\cite{kld}, focal loss $\mathcal{L}_{focal}$~\cite{deim}, and L1 loss $\mathcal{L}_{L1}$, while O$^2$-DFINE further incorporates an additional Fine-Grained Localization (FGL) loss $\mathcal{L}_{fgl}$~\cite{dfine}. The weights of losses are $\lambda_{kld}=2.0$, $\lambda_{cls}=1.0$, $\lambda_{L1}=5.0$, and $\lambda_{fgl}=0.15$ respectively. Overall loss is calculated as $\lambda_{kld}\mathcal{L}_{kld}+\lambda_{cls}\mathcal{L}_{focal}+\lambda_{L1}\mathcal{L}_{L1}+\lambda_{fgl}\mathcal{L}_{fgl}$.  The distillation loss is not used. The weights of cost are $\xi_{kld}=2.0$, $\xi_{cls}=2.0$, and $\xi_{cf}=5.0$ for KLD cost $\mathcal{C}_{kld}$, focal cost $\mathcal{C}_{focal}$, and Chamfer distance cost $\mathcal{C}_{cf}$. Overall cost is calculated as $\xi_{kld}\mathcal{C}_{kld}+\xi_{cls}\mathcal{C}_{focal}+\xi_{cf}\mathcal{C}_{cf}$. Details of experiments are displayed in Table~\ref{expsetting}.

For experiments on the DOTA and FAIR1M, images are cropped into 1024$\times$1024 patches with an overlap of 200. For multi-scale training on DOTA-v1.0, each image is first resized to three scales (0.5, 1.0, and 1.5) and then cropped following single-scale training. For DIOR-R, images are used at the original fixed size of 800$\times$800. For DOTA and FAIR1M, we submit test predictions to their official evaluation servers to obtain the final performance. DOTA uses both random flips and random rotations for data augmentation, while DIOR-R and FAIR1M apply only random flips.

\textit{Evaluation metrics.}
Average Precision (AP) evaluates detection accuracy by integrating the precision–recall curve over different confidence thresholds. The parameter count measures the number of learnable weights in the model. FLOPs represent the total number of floating-point operations required for a single forward pass. FPS reflects the inference speed by reporting how many images are processed per second. Latency is the time required to process one image, and it is inversely related to FPS.

\subsection{Comparisons with modern oriented object detector}

\DIORRResults

\TableDOTAVonefiveResult

\textit{Results on DOTA-v1.0.}
Table \ref{DOTA-1.0-result} summarizes the experimental results on the DOTA-v1.0. O$^2$-DFINE-s / m / l achieve AP$_{50}$ of 76.14\%, 77.61\%, and 77.73\%, respectively, consistently outperforming the corresponding DFINE-R baselines. Similarly, O$^2$-RTDETR with R18 / R34 / R50 backbones attains AP$_{50}$ of 77.31\%, 78.13\%, and 78.45\%, respectively, surpassing the RTDETR-R models. Furthermore, under multi-scale training and testing, O$^2$-DEIM-R18 / R34 / R50 achieve AP$_{50}$ of 79.49\%, 80.04\%, and 80.15\%, respectively. The compared methods use the same image preprocessing as this chapter. These results demonstrate that our method is highly competitive with both modern non-end-to-end real-time and non-real-time oriented object detectors.

\textit{Comparison of FPS, Parameters, FLOPs and Accuracy.}
Table \ref{DOTA-1.0-result} shows the comparison in terms of FPS, parameters, FLOPs, and accuracy. All FLOPs are computed with a 1024$\times$1024 input, and FPS is measured on an NVIDIA \textcolor[HTML]{76B900}{\raisebox{-0.15\height}{\simpleicon{nvidia}}} 2080ti using TensorRT FP16 under the same resolution. O$^2$-DFINE-s reaches 297 FPS with only 10M parameters and 60G FLOPs, while the larger O$^2$-DFINE-m/l variants run at 176/132 FPS with 19M/31M parameters and 142G/229G FLOPs. O$^2$-RTDETR with ResNet-18/34/50 backbones achieves 233/174/119 FPS, using 20M/31M/42M parameters and 147G/228G/339G FLOPs, respectively. The O$^2$-DEIM retains the same FPS, parameter count, and FLOPs as its baseline. Importantly, our approach introduces no additional parameters, latency, or FLOPs, as it focuses on representation and optimization. Our approach is competitive with existing real-time detectors in both efficiency and accuracy.

\TableFAIRResult

\textit{Results on DIOR-R.}
The DIOR-R results are summarized in Table \ref{DIOR-R-result}. Our method achieves consistently strong performance across all model variants. O$^2$-DFINE-s attains 67.05\% AP$_{50}$, while the larger O$^2$-DFINE-m further improves the score to 69.86\% AP$_{50}$. For the RTDETR family, O$^2$-RTDETR-R34 obtains 68.67\% AP$_{50}$, and the stronger O$^2$-RTDETR-R50 reaches 72.26\% AP$_{50}$, outperforming all existing oriented DETR-based and two-stage detectors listed in the comparison. These results demonstrate that our oriented modeling brings consistent gains across backbones and model scales.
% and achieves state-of-the-art performance on DIOR-R.

\textit{Results on DOTA-v1.5.}
As shown in Table \ref{DOTA-1.5-result}, our proposed models achieve competitive performance on the DOTA-v1.5 dataset. Specifically, O$^2$-DFINE-s attains 70.22\% AP$_{50}$, while O$^2$-DFINE-m reaches 72.15\% AP$_{50}$. Furthermore, O$^2$-RTDETR-R34 achieves 71.91\% AP$_{50}$, and O$^2$-RTDETR-R50 attains 73.76\% AP$_{50}$, outperforming several existing approaches with particularly significant gains on small objects such as helicopters and container cranes.

\textit{Results on FAIR1M-v1.0.}  
Table~\ref{FAIR1M-1.0-result} presents the performance comparison on the FAIR1M-v1.0 dataset under single-scale training and testing settings. Our methods achieve strong results across the board: O$^2$-DFINE-s obtains an AP$_{50}$ of 39.77\%, and its larger variant O$^2$-DFINE-m further improves this to 42.01\%. Moreover, O$^2$-RTDETR with a ResNet-34 backbone achieves 40.45\% AP$_{50}$, while the ResNet-50 version reaches 43.14\% AP$_{50}$, outperforming all existing approaches by a clear margin.

\EachProposedModuleOnDOTA

\AblationAngleDistriRefine

\AblationCost

\subsection{Ablation Study}

\textit{Model Improvement Roadmap.}
Table \ref{EachProposedModuleOnDOTA} presents a step-by-step ablation study illustrating how each component contributes to the final performance on DOTA-v1.0. For O$^2$-DFINE-R18, AP$_{50}$ increases from 72.44\% to 76.14\% as oriented contrastive denoising, angle distribution refinement, and the Chamfer distance cost are introduced. For O$^2$-RTDETR-s, these components similarly raise AP$_{50}$ from 73.94\% to 77.31\%. Adding Matchability-Aware Loss~\cite{deim} and multi-scale training further improves O$^2$-DEIM-s to 79.49\%. Following ~\cite{deim}, multi-scale training as data augmentation provides more positive objects. All improvements come without extra cost, as our method focuses on detection optimization rather than modifying the feature extraction module.

\textit{Number of bins $N$}.
Table \ref{ablation_angle_distri_refine} evaluates the effect of the number of angle bins in the Angle Distribution Refinement. Increasing $N$ from 8 to 32 gradually improves performance, with the best result of 76.14\% AP$_{50}$ achieved at $N=32$. Further increasing the bin count to 64 or 128 does not provide additional gains and even leads to slight fluctuations.

\textit{Weighting function parameters $a,c$.} We examine the hyper-parameters $a,c$ in weighting function $\mathcal{A}(\cdot)$ in Table \ref{ablation_angle_distri_refine}. When $c$ becomes excessively large, the weighting function degenerates toward a uniformly spaced linear form, which leads to suboptimal AP performance. Moreover, too large or small values of $a$ change the bounds of $\mathcal{A}(\cdot)$, thereby degrading localization accuracy.

\textit{Different distance cost.}
Table \ref{ablation_cost} compares different distance costs for bipartite matching. Using only KL divergence provides the weakest supervision, resulting in 73.69\% AP$_{50}$. In the subsequent experiments, KL divergence is kept as the IoU cost while additional distance terms are introduced. L1 distance yields a moderate improvement to 74.28\%, and adding geometric information through the Hausdorff distance further boosts accuracy to 75.05\%. The Chamfer distance achieves the best performance of 76.14\%, showing that it captures oriented object geometry more accurately.

\textit{Chamfer distance cost weight.}
We further examine the effect of the Chamfer distance weight in Table \ref{ablation_cost}. A balanced weight of 5.0 yields the highest 76.14\% AP$_{50}$, whereas smaller weights such as 4.0 and larger weights such as 6.0 lead to slight accuracy degradation. This suggests that unsuitable geometric constraints can disrupt the matching quality.

\visualContrastiveDenoise

\AblationDenoise

\textit{Point number in Chamfer distance cost.}
The number of points used in the Chamfer distance is also evaluated in Table \ref{ablation_cost}. Using the four vertices $\{v1,v2,v3,v4\}$ of the oriented box gives the best performance of 76.14\% AP$_{50}$. When more points are uniformly sampled along the four edges to form 16 or 32 points, the accuracy slightly decreases. This is because the four vertices already represent the convex hull of the box, and adding edge points increases the size of the point sets for both predictions and ground truths, causing small fluctuations in the Chamfer distance.

\textit{Analysis of oriented contrastive denoising.}
Table \ref{ablation_denoise} analyzes box noise, angle noise, geometric noise, and probability noise in oriented contrastive denoising. 
Box noise achieves the best performance, reaching 76.14\% AP$_{50}$ with $\lambda_1=1.0, \lambda_2=2.0$. 
In contrast, angle noise perturbs only the rotation $\theta$ while keeping $(c_x,c_y,w,h)$ unchanged, thus producing a much narrower range of variations.
Geometric noise, which combines both box noise and angle noise, tends to generate perturbed boxes that deviate excessively from the ground truth. Since oriented boxes are sensitive to rotation, changes in $\theta$ can lead to IoU drops, reducing their effectiveness.
Probability noise perturbs the covariance representation of the box, but the induced changes are relatively small and do not effectively influence the box center $(c_x,c_y)$, resulting in weaker supervision.

\visualChamfer

When no noise is added, both positive and negative samples are the same as ground-truth boxes. This case leads to two issues. First, the positive samples perfectly match the ground-truth boxes, resulting in zero loss. Second, identical samples are assigned as foreground objects and background at the same time, which confuses the model during training.

\textit{Number of denoising query.}
Table \ref{ablation_denoise} evaluates the impact of the number of denoising queries. We test a total of 100, 200, and 300 denoising queries, where positive and negative queries are kept in a 1:1 ratio. Increasing the query count from 100 to 200 improves AP$_{50}$ from 75.83\% to 76.14\%, while increasing the number to 300 yields nearly a similar result. When the number of denoising queries is small, the randomly generated noised boxes cover only limited cases. As the number increases, the model benefits from richer noise diversity until it saturates.

\subsection{Qualitative Analysis}

\textit{Oriented contrastive denoising.}
\figurename~\ref{visual_contrastive_denoising} illustrates examples of oriented contrastive denoising under different noise types. The figure includes a box noise example with $\lambda_1=1.0$ and $\lambda_2=2.0$, an angle noise example with $\lambda_3=9.0$ and $\lambda_4=18.0$, a geometric noise example combining box and angle perturbations with $\lambda_1=1.0, \lambda_2=2.0, \lambda_3=9.0, \lambda_4=18.0$, and a probability noise example with $\lambda_5=0.3$ and $\lambda_6=0.6$. This provides intuitive insight into how different noise formulations perturb oriented boxes during training.

\textit{Chamfer distance cost vs. other costs.} \figurename~\ref{visual_chamfer} displays examples comparing the Chamfer distance cost with KL divergence, L1, and Hausdorff distance costs. The KL divergence cost is zero when two squares have the same size and center. The L1 distance cost assigns a box far from the ground truth rather than a nearby box. The Hausdorff distance cost is determined by the farthest distance between two point sets. Our Chamfer distance cost effectively overcomes these limitations.

\visualAngleDistRefine

\heatmap

\visualRoatedAttn

\textit{Angle distribution refinement.}
The angle distribution refinement analysis is shown in \figurename~\ref{visual_angle_dist_refine}. The golden external rectangles and vertex offsets represent initial predictions from the first decoder layer, while the red ones denote refined predictions from the final layer. The golden curves represent the initial unweighted and weighted probability distributions for the vertex offsets $(\epsilon,\eta)$ and the external rectangle edges $(\alpha, \beta, \gamma, \delta)$, while the red curves represent the refined distributions. In contrast to DFL~\cite{ppyoloer} without progressive angle refinement, our approach captures this process explicitly. The initial boxes roughly locate the target objects, and the refined distribution further fine-tunes the localization.

\textit{Feature response analysis.}
\figurename~\ref{visual_heatmap} visualizes the backbone output feature maps using heatmaps. 
Compared with other methods, our method produces more compact and discriminative feature responses that align better with object regions. High activations concentrate on object areas, while background responses are suppressed. This demonstrates an improved geometry-aware feature representation for oriented detection.

\textit{Sampling points in cross attention.}
In vanilla cross attention, sampling points are axis-aligned and do not follow object orientation, leading to attention regions misaligned with object boundaries, as shown in \figurename~\ref{visual_rotated_attn}(a). In contrast, our method rotates sampling points according to object angles, allowing them to better align with oriented objects.

\textit{Result comparison with other detectors.} \figurename~\ref{visual_results} illustrates a qualitative comparison between our approach and existing detectors on the DOTA-v1.0 dataset across challenging scenarios, including large-scale objects, dense object distributions, and low-light imagery. Our method outperforms other approaches on large objects, low-light scenes, and dense object layouts, yielding more accurate and stable results. 
\figurename~\ref{visual_rotated_attn}(b) shows suboptimal detection results in dense scenes. Since detection transformers use a fixed number of queries, missed detections occur when the number of objects exceeds the query count.

\visualresults

\section{Conclusion}

In this paper, we present the first real-time end-to-end oriented object detection transformer for remote sensing images. We first proposed Angle Distribution Refinement, which reformulates angle regression as an iterative refinement of probability distributions to capture rotation uncertainty. To mitigate geometric ambiguity during label assignment, we further introduced the Chamfer distance matching cost. In addition, we designed Oriented Contrastive Denoising with four noise modes to stabilize training, and proposed an instability metric to analyze assignment consistency across decoder layers. Based on these designs, we developed a family of real-time oriented detectors, including O$^2$-RTDETR, O$^2$-DFINE, and O$^2$-DEIM. Extensive experiments on four remote sensing benchmarks demonstrate that our approach achieves a favorable balance between accuracy and speed.  Code is available at \url{https://github.com/wokaikaixinxin/ai4rs}.

\appendix

\textit{Rotation-Aware Cross-Attention.}
In O$^2$-RTDETR and O$^2$-DEIM cross-attention, the sampling points are rotated around the box center according to the predicted box angle. 
In D-FINE~\cite{dfine}, the cross-attention performs uneven sampling across multiple feature scales. We extend this idea by rotating these unevenly sampled points according to the box orientation, forming the cross-attention used in O$^2$-DFINE.

\textit{Sampling points in cross-attention.}
\figurename~\ref{appendix_visual_rotated_attn} visualizes the sampling points in cross-attention. In Rotated Deformable DETR, sampling points are axis-aligned and do not follow object orientation, leading to misaligned attention regions. In contrast, O$^2$-RTDETR and O$^2$-DFINE rotate sampling points according to object angles. This allows sampling points to better align with oriented objects.

\AppendixvisualRoatedAttn

\bibliographystyle{IEEEtran}
\bibliography{reference}

\end{document}